\pdfoutput=1
\documentclass{article}

\usepackage[final]{neurips_2024}

\usepackage[utf8]{inputenc} %
\usepackage[T1]{fontenc}    %
\usepackage{url}            %
\usepackage{booktabs}       %
\usepackage{amsfonts}       %
\usepackage{nicefrac}       %
\usepackage{microtype}      %
\usepackage{xcolor}         %
\usepackage[colorlinks,citecolor=DarkGreen,linkcolor=FireBrick,urlcolor=FireBrick,linktocpage=true,unicode]{hyperref}
\usepackage[utf8]{inputenc}
\usepackage{amsmath}
\usepackage{amssymb}
\usepackage{mathtools}
\numberwithin{equation}{section}
\usepackage{slashed}
\usepackage{braket}
\usepackage{enumitem}

\allowdisplaybreaks
\usepackage[colorinlistoftodos,prependcaption,textsize=tiny]{todonotes}
\usepackage{xargs}                      %

\usepackage{graphicx}
\usepackage{tikz}
\usepackage{tikz-cd}
\usepackage{times}
\usepackage{courier}
\usepackage{bm}
\usepackage{physics}
\usepackage{xcolor}
\usepackage{natbib}
\usepackage{mdframed}
\usepackage{nicefrac}
\usepackage{booktabs}
\usepackage{lipsum}
\usepackage{titlesec}
\usepackage{wrapfig,lipsum,booktabs}
\usepackage{authblk}
\usepackage{blindtext}
\usepackage{titletoc}
\usepackage{multirow} 
\usepackage{algorithm}
\usepackage{algorithmic}
\usepackage{titletoc}

\newcommandx{\unsure}[2][1=]{ \todo[linecolor=red,backgroundcolor=red!25,bordercolor=red, #1]{Unsure: #2}}
\newcommandx{\change}[2][1=]{\todo[linecolor=blue,backgroundcolor=blue!25,bordercolor=blue,#1]{Change: #2}}
\newcommandx{\toadd}[2][1=]{\todo[linecolor=teal,backgroundcolor=teal!25,bordercolor=teal,#1]{Add: #2}}
\newcommandx{\improvement}[2][1=]{\todo[linecolor=Plum,backgroundcolor=Plum!25,bordercolor=Plum,#1]{Improve: #2}}
\newcommandx{\thiswillnotshow}[2][1=]{\todo[disable,#1]{#2}}
\usepackage{pifont}%

\usepackage{authblk}
\definecolor{DarkGreen}{rgb}{0,0.40,0}
\definecolor{FireBrick}{rgb}{0.698,0.133,0.133}
\usepackage{colortbl}

\usepackage{float}
\usepackage{xcolor}
\definecolor{LightCyan}{rgb}{0.8, 0.9, 1}

\newcolumntype{g}{>{\columncolor{LightCyan}\hspace{0pt}}c}

\usepackage[nameinlink,capitalize,noabbrev]{cleveref}

\usepackage{CJK}

\usepackage{array}
\usepackage{makecell}

\def\half{{\frac{1}{2}}}

\newcommand{\bea}{\begin{eqnarray}}
\newcommand{\eea}{\end{eqnarray}}

\usepackage{slashed}

\def\({\left(}
\def\){\right)}
\def\[{\left[}
\def\]{\right]}

\usepackage{dashbox}
\usepackage{xcolor}
\usepackage{colortbl}
\usepackage{arydshln}
\definecolor{lightyellow}{rgb}{1.0, 0.95, 0.7}
\definecolor{blue}{rgb}{0.0, 0.4, 1.0}
\definecolor{Blue}{rgb}{0,0,1}
\definecolor{darkgreen}{rgb}{0,0.40,0}
\definecolor{firebrick}{rgb}{0.698,0.133,0.133}

\definecolor{colorA}{rgb}{1,0,0}
\definecolor{colorB}{rgb}{0,0.3,1}
\definecolor{colorC}{rgb}{0.9,0.8,0.2}
\definecolor{colorD}{rgb}{0,0.65,0}
\definecolor{lesslightgray}{rgb}{0.5,0.5,0.5}
\definecolor{light-gray}{gray}{0.95}

\let\tilde\widetilde

\newcommand{\calC}{\mathcal{C}}

\newcommand{\calH}{\mathcal{H}}

\newcommand{\calK}{\mathcal{K}}
\newcommand{\calL}{\mathcal{L}}

\newcommand{\calO}{\mathcal{O}}

\newcommand{\calT}{\mathcal{T}}

\newcommand{\calV}{\mathcal{V}}

\newcommand{\bC}{C}

\newcommand{\bW}{W}
\newcommand{\bX}{X}

\newcommand{\ba}{a}
\newcommand{\bb}{b}
\newcommand{\bc}{c}

\newcommand{\bp}{p}

\newcommand{\bv}{v}

\newcommand{\bx}{x}
\newcommand{\by}{y}
\newcommand{\bz}{z}
\newcommand{\bxi}{\xi}

\newcommand{\Min}{\mathop{\rm Min}}
\newcommand{\argmin}{\mathop{\mathrm{argmin}}}

\newcommand{\Softmax}{\mathop{\rm{Softmax}}}

\newcommand{\lse}{\mathop{\rm{lse}}}

\newcommand{\sT}{ \mathsf{T} }

\def\R{\mathbb{R}}

\let\cite\citep 

\usepackage{amsthm}
\makeatletter
\def\th@remark{%
  \thm@headfont{\bfseries}%
  \normalfont %
  \thm@preskip\topsep \divide\thm@preskip\tw@
  \thm@postskip\thm@preskip
}
\makeatother
\usepackage[many]{tcolorbox}
\theoremstyle{definition}
\newtheorem{theorem}{Theorem}[section]
\tcolorboxenvironment{theorem}{
  breakable,
  colback=black!10,
  colframe=white,%
  width=\dimexpr\linewidth+10pt\relax,%
  enlarge left by=-5pt,%
  enlarge right by=-5pt,%
  boxsep=5pt,%
  boxrule=0pt,
  left=0pt,right=0pt,top=0pt,bottom=0pt,
  sharp corners,
  before skip=\topsep,
  after skip=\topsep
}
\newtheorem{lemma}{Lemma}[section]
\tcolorboxenvironment{lemma}{
  breakable,
  colback=black!10,
  colframe=white,%
  width=\dimexpr\linewidth+10pt\relax,%
  enlarge left by=-5pt,%
  enlarge right by=-5pt,%
  boxsep=5pt,%
  boxrule=0pt,
  left=0pt,right=0pt,top=0pt,bottom=0pt,
  sharp corners,
  before skip=\topsep,
  after skip=\topsep
}

\tcolorboxenvironment{corollary}{
  breakable,
  colback=black!10,
  colframe=white,%
  width=\dimexpr\linewidth+10pt\relax,%
  enlarge left by=-5pt,%
  enlarge right by=-5pt,%
  boxsep=5pt,%
  boxrule=0pt,
  left=0pt,right=0pt,top=0pt,bottom=0pt,
  sharp corners,
  before skip=\topsep,
  after skip=\topsep
}

\newtheorem{assumption}{Assumption}
\tcolorboxenvironment{assumption}{
  breakable,
  colback=black!10,
  colframe=white,%
  width=\dimexpr\linewidth+10pt\relax,%
  enlarge left by=-5pt,%
  enlarge right by=-5pt,%
  boxsep=5pt,%
  boxrule=0pt,
  left=0pt,right=0pt,top=0pt,bottom=0pt,
  sharp corners,
  before skip=\topsep,
  after skip=\topsep}

\newtheorem{proposition}{Proposition}[section]
\tcolorboxenvironment{proposition}{
  breakable,
  colback=black!10,
  colframe=white,%
  width=\dimexpr\linewidth+10pt\relax,%
  enlarge left by=-5pt,%
  enlarge right by=-5pt,%
  boxsep=5pt,%
  boxrule=0pt,
  left=0pt,right=0pt,top=0pt,bottom=0pt,
  sharp corners,
  before skip=\topsep,
  after skip=\topsep
}
\theoremstyle{definition}
\newtheorem{definition}{Definition}[section]
\tcolorboxenvironment{definition}{
  breakable,
  colback=black!10,
  colframe=white,%
  width=\dimexpr\linewidth+10pt\relax,%
  enlarge left by=-5pt,%
  enlarge right by=-5pt,%
  boxsep=5pt,%
  boxrule=0pt,
  left=0pt,right=0pt,top=0pt,bottom=0pt,
  sharp corners,
  before skip=\topsep,
  after skip=\topsep
}
\theoremstyle{remark}
\newtheorem{remark}{Remark}

\newtheorem{problem}{Problem}
\tcolorboxenvironment{problem}{
  breakable,
  colback=black!10,
  colframe=white,%
  width=\dimexpr\linewidth+10pt\relax,%
  enlarge left by=-5pt,%
  enlarge right by=-5pt,%
  boxsep=5pt,%
  boxrule=0pt,
  left=0pt,right=0pt,top=0pt,bottom=0pt,
  sharp corners,
  before skip=\topsep,
  after skip=\topsep
}

\tcolorboxenvironment{question}{
  breakable,
  colback=black!10,
  colframe=white,%
  width=\dimexpr\linewidth+10pt\relax,%
  enlarge left by=-5pt,%
  enlarge right by=-5pt,%
  boxsep=5pt,%
  boxrule=0pt,
  left=0pt,right=0pt,top=0pt,bottom=0pt,
  sharp corners,
  before skip=\topsep,
  after skip=\topsep
}
\crefname{question}{Question}{Questions}

\crefname{theorem}{Theorem}{Theorems}
\crefname{proposition}{Proposition}{Propositions}
\crefname{lemma}{Lemma}{Lemmas}
\crefname{corollary}{Corollary}{Corollaries}
\crefname{definition}{Definition}{Definitions}
\crefname{assumption}{Assumption}{Assumptions}
\crefname{remark}{Remark}{Remarks}
\crefname{problem}{Problem}{Problems}
\crefname{property}{Property}{property}

\numberwithin{equation}{section}
\numberwithin{theorem}{section}
\numberwithin{proposition}{section}
\numberwithin{lemma}{section}

\usepackage{lipsum}

\newcommand*{\annot}[1]{\tag*{\footnotesize{\textcolor{black!50}{\big(#1\big)}}}}

\makeatletter
\let\save@mathaccent\mathaccent
\newcommand*\if@single[3]{%
    \setbox0\hbox{${\mathaccent"0362{#1}}^H$}%
    \setbox2\hbox{${\mathaccent"0362{\kern0pt#1}}^H$}%
    \ifdim\ht0=\ht2 #3\else #2\fi
}
\newcommand*\rel@kern[1]{\kern#1\dimexpr\macc@kerna}
\newcommand*\widebar[1]{\@ifnextchar^{{\wide@bar{#1}{0}}}{\wide@bar{#1}{1}}}
\newcommand*\wide@bar[2]{\if@single{#1}{\wide@bar@{#1}{#2}{1}}{\wide@bar@{#1}{#2}{2}}}
\newcommand*\wide@bar@[3]{%
    \begingroup
    \def\mathaccent##1##2{%
        \let\mathaccent\save@mathaccent
        \if#32 \let\macc@nucleus\first@char \fi
        \setbox\z@\hbox{$\macc@style{\macc@nucleus}_{}$}%
        \setbox\tw@\hbox{$\macc@style{\macc@nucleus}{}_{}$}%
        \dimen@\wd\tw@
        \advance\dimen@-\wd\z@
        \divide\dimen@ 3
        \@tempdima\wd\tw@
        \advance\@tempdima-\scriptspace
        \divide\@tempdima 10
        \advance\dimen@-\@tempdima
        \ifdim\dimen@>\z@ \dimen@0pt\fi
        \rel@kern{0.6}\kern-\dimen@
        \if#31
        \overline{\rel@kern{-0.6}\kern\dimen@\macc@nucleus\rel@kern{0.4}\kern\dimen@}%
        \advance\dimen@0.4\dimexpr\macc@kerna
        \let\final@kern#2%
        \ifdim\dimen@<\z@ \let\final@kern1\fi
        \if\final@kern1 \kern-\dimen@\fi
        \else
        \overline{\rel@kern{-0.6}\kern\dimen@#1}%
        \fi
    }%
    \macc@depth\@ne
    \let\math@bgroup\@empty \let\math@egroup\macc@set@skewchar
    \mathsurround\z@ \frozen@everymath{\mathgroup\macc@group\relax}%
    \macc@set@skewchar\relax
    \let\mathaccentV\macc@nested@a
    \if#31
    \macc@nested@a\relax111{#1}%
    \else
    \def\gobble@till@marker##1\endmarker{}%
    \futurelet\first@char\gobble@till@marker#1\endmarker
    \ifcat\noexpand\first@char A\else
    \def\first@char{}%
    \fi
    \macc@nested@a\relax111{\first@char}%
    \fi
    \endgroup
    }
\makeatother

\makeatletter
\newcommand*{\redefinesymbolwitharg}[1]{%
  \expandafter\let\csname ltx#1\expandafter\endcsname\csname #1\endcsname
  \@namedef{#1}{\@ifnextchar{^}{\@nameuse{#1@}}{\@nameuse{#1@}^{}}}%
  \expandafter\def\csname #1@\endcsname^##1##2{%
     \csname ltx#1\endcsname\ifx!##1!\else^{##1}\fi\mathopen{}\mathclose\bgroup\left(##2\aftergroup\egroup\right)
     }%
}
\makeatother
\redefinesymbolwitharg{sin}
\redefinesymbolwitharg{cos}

\newcommand{\uhop}{{$\mathtt{U\text{-}Hop}$}}

\titlespacing\section{0pt}{4pt plus 4pt minus 2pt}{-2pt plus 2pt minus 2pt}
\titlespacing\subsection{0pt}{2pt plus 4pt minus 2pt}{-2pt plus 2pt minus 2pt}
\titlespacing\subsubsection{0pt}{2pt plus 4pt minus 2pt}{-2pt plus 2pt minus 2pt}

\setlist[itemize]{leftmargin=1em, before=\vspace{-0.5em}, after=\vspace{-0.5em}, itemsep=0.1em}
\setlist[enumerate]{leftmargin=1.2em, before=\vspace{-0.5em}, after=\vspace{-0.5em}, itemsep=0.1em}

\title{Provably Optimal Memory Capacity for 
Modern Hopfield Models: 
 Transformer-Compatible 
 Dense Associative Memories as Spherical Codes

}

\author{
{\bf 
Jerry Yao-Chieh Hu\thanks{Equal contribution. 
Code is available
at \href{https://github.com/MAGICS-LAB/Optimal-Hopfield-Memory}{GitHub}. 
Latest version is on \href{https://arxiv.org/abs/2410.23126}{arXiv}.
}\;\,$^{\dagger\ddag}$
\quad
Dennis Wu$^{*\ddag}$
\quad
Han Liu$^{\dagger\ddag\S}$}
\\
{\small
    $^\dagger$Center for Foundation Models and Generative AI, $^\ddag$Department of Computer Science, $^\S$Department of Statistics and Data Science, Northwestern University, Evanston, IL 60208, USA
   \texttt{\{\href{mailto:jhu@u.northwestern.edu}{jhu},\href{mailto:hibb@u.northwestern.edu}{hibb}\}@u.northwestern.edu,
   \href{mailto:hanliu@northwestern.edu}{hanliu@northwestern.edu}
   }
   }
   \vspace{-1.5em}
}

\titlespacing\section{0pt}{4pt plus 4pt minus 2pt}{-2pt plus 2pt minus 2pt}
\titlespacing\subsection{0pt}{2pt plus 4pt minus 2pt}{-2pt plus 2pt minus 2pt}
\titlespacing\subsubsection{0pt}{2pt plus 4pt minus 2pt}{-2pt plus 2pt minus 2pt}
\titlespacing\paragraph{0pt}{2pt plus 4pt minus 2pt}{1em}

\usepackage{titletoc}

\begin{document}

\maketitle

\begin{abstract}

We study the optimal memorization capacity of modern Hopfield models and Kernelized Hopfield Models (KHMs), a transformer-compatible class of Dense Associative Memories.
We present a tight analysis by establishing a connection between the memory configuration of KHMs and spherical codes from information theory. 
Specifically, we treat the stored memory set as a specialized spherical code.
This enables us to cast the memorization problem in KHMs into a point arrangement problem on a hypersphere.
We show that the optimal capacity of KHMs occurs when the feature space allows memories to form an optimal spherical code.
This unique perspective leads to: (i) An analysis of how KHMs achieve optimal memory capacity, and identify corresponding necessary conditions. 
Importantly, we establish an upper capacity bound that matches the well-known exponential lower bound in the literature. 
This provides the first tight and optimal asymptotic memory capacity  for modern Hopfield models.
(ii) A sub-linear time algorithm $\mathtt{U}\text{-}\mathtt{Hop}$+ to reach KHMs' optimal capacity. 
(iii) An analysis of the scaling behavior of the required feature dimension relative to the number of stored memories.
These efforts improve both the retrieval capability of KHMs and the representation learning of corresponding transformers.
Experimentally, we provide thorough numerical results to back up theoretical findings.

\end{abstract}

\section{Introduction}
\label{sec:intro}
We study the optimal memorization capacity of 
\underline{K}ernelized modern \underline{H}opfield \underline{M}odels (KHMs) \cite{wu2024uniform}, propose a sublinear-time algorithm to achieve it, and analyze parameter selection  for these models.
KHMs belong to a class of \textit{transformer-compatible} Dense Associative Memory \cite{krotov2020large,krotov2016dense} known as Modern Hopfield Models (MHMs) \cite{wu2024uniform,wu2023stanhop,hu2024outlier,hu2023SparseHopfield,ramsauer2020hopfield}. The defining characteristics of these models include their super-linear memory capacity and strong connection to transformer attention mechanisms \cite{vaswani2017attention}.
The former makes them interesting models for associative memory, and the latter 
makes them versatile transformer-compatible backbones with diverse empirical successes \cite{burns2024semantically,burns2023simplicial,hu2024outlier,hu2024computational,xu2024bishop,wu2024uniform,wu2023stanhop,hoover2023energy,seidl2022improving,furst2022cloob}.
However,
one major limitation of MHMs is their reliance on the quality of memory distribution for effective pattern storage and retrieval \cite[Sec.~1]{wu2024uniform}.

Studying this limitation in these models is fundamental and of practical importance. 
One one hand, it prevents MHMs from functioning as full-fledged content-addressable memory models.
On the other hand, it implies that the representation learning ability of current transformer attention \cite{vaswani2017attention} is suboptimal \cite[Thm.~3.1]{wu2024uniform}.
Addressing this issue benefits both computational neuroscience and large foundation model research \cite{bietti2024birth,krotov2023new,kozachkov2023building,cabannes2023scaling,hoover2023memory}.
\underline{K}ernelized modern \underline{H}opfield \underline{M}odels (KHMs) \cite{wu2024uniform} alleviate this issue by storing memories in the kernelized feature space. 
A key advantage of KHMs is their ability to reposition memories in the feature space, resulting in larger storage capacity.
However, despite strong empirical performance, their capacity still lacks an optimal guarantee \cite[Sec.~5]{wu2024uniform}.
In this work, we close this gap by establishing the optimality of KHMs' memory capacity and presenting a sublinear-time algorithm to achieve it.

Let $\Xi \coloneqq  [ \bxi_1,\cdots,\bxi_M ] \in \R^{d\times M}$ be a set of memory patterns where each column (indexed by $\mu \in [M]$) represents a memory $\bxi_\mu\in\R^{d}$, and let $\bx\in\R^d$ be the input query.
The Hopfield models \cite{hopfield1982neural} are energy-based associative memory models, which store memories on the local minima of their energy functions.
They retrieve a pattern by iteratively updating the query $ \bx^t \mapsto \bx^{t+1}$ with its update rule $\calT(\bx^t)$, for some $t \in \{ 0, 1, ... \}$.
This update rule converge to a fixed
point $\bx^\star$, defined by $\bx^\star = \calT(\bx^\star)$.
$\bx^\star$ is the retrieved pattern\footnote{$x^\star$ corresponds either to one of the memories or to a \textit{fuzzy} combination of them.
Please see \citep[Sec.~A.1.5]{ramsauer2020hopfield} for an informal discussion, \citep[Sec. 1]{wu2024uniform} for further discussion, and \citep{santos2024sparse,wu2024uniform} for conditions of exact retrieval through sparsity \cite{wu2023stanhop,hu2023SparseHopfield}.
} based on initial query $x^0$.

Explicitly, iteratively updating $\bx$ with $\calT$ is defined as a process of minimization to an energy function $E(\bx)$.
For example, the Modern Hopfield Model \cite{ramsauer2020hopfield} has the energy function:
\begin{align}\label{MHM-energy}
    E(\bx) 
    = 
    \frac{1}{2} \Braket{ \bx, \bx } +
    \lse \(  \beta, \Xi^\sT \bx \),
\end{align}
where $\lse(\beta, \bz) \coloneqq \log( \sum_{\mu=1}^M \exp{ \beta z_\mu }) $,
with some $\beta > 0$. 
With the Concave-Convex Procedure (CCCP) \cite{yuille2001concave}, \eqref{MHM-energy} is monotonically decreased by an iterative update rule
\begin{equation}\label{MHM-update}
   \bx^{t+1} \leftarrow \calT(x^t) = \Xi \cdot \Softmax ( \beta \Xi^\sT \bx^t ).
\end{equation}
This design boosts MHMs to store \textit{exponentially} (in pattern dimension $d$) many memories compared to the \textit{linear} capacity of the classic Hopfield model \cite{hopfield1982neural}.
It also provides a model-based interpretation of the transformer attention mechanism \cite{wu2024uniform,wu2023stanhop,hu2024outlier,hu2024nonparametric,hu2023SparseHopfield,ramsauer2020hopfield}.
However, their retrieval accuracy and memory capacity hinge on the quality of the stored memory set \citep[Sec.~1]{wu2024uniform}, and hence are suboptimal in most scenarios.

To be concrete, for retrieving the $\mu$-th memory ($\mu\in[M]$), the retrieval error of MHM is exponentially suppressed by the pattern separation: $\Delta_\mu\coloneqq \Braket{ \bxi_\mu, \bxi_\mu } - \max_{\nu,\nu\neq\mu} \Braket{\bxi_\nu, \bxi_\mu}$ (\cite[Eqn.~1.3]{wu2024uniform} or \cite[Eqn.~2.7]{hu2023SparseHopfield}).
This $\Delta_\mu$-dependence in MHM retrieval accuracy also manifests the $\Delta_\mu$-dependence in memory capacity. 
To see this, recall that the standard memory capacity is a high-probability bound based on thresholding the separation $\Delta_\mu$ for each pattern $\mu \in [M]$ to determine storage and retrieval (\cref{high-prob-bound}). 
Explicitly, storing a pattern requires its separation to exceed a threshold that decreases with the \textit{minimal} separation: $\Delta_{\min} \coloneqq \min_{\mu \in [M]} \Delta_\mu$.
Namely, a larger $\Delta_{\min}$ leads to larger capacity 
(\cref{appendix:MHM-capacity-explained}).
Thus, the capacity depends on $\Delta_{\min}$.
Yet, $\Delta_{\min}$ depends on the stored memories $\Xi$.
This $\Xi$-dependence makes the capacity suboptimal.

\citet{wu2024uniform} relax such limitation 
by introducing a kernel as a learnable similarity measure, using stored memory patterns as training data to enhance memory capacity.
Specifically, they propose the Kernelized Modern Hopfield Model (KHM) defined by following update rule and energy function:
\begin{align}   
\label{KHM-update-rule}
   \bx^{t+1} 
   \leftarrow \calT_\Phi( \bx^t )
   \coloneqq
   \Xi
   \cdot
   \Softmax
   \(
   \beta
   \calK( \Xi, \bx )
   \),
    \quad
    E_\calK(\bx) 
    = 
    \half \calK(\bx,\bx) +
    \lse \( \beta, \calK( \Xi, \bx ) \),
\end{align}
where the kernel $\calK(\cdot,\cdot)\coloneqq \Braket{\Phi(\cdot),\Phi(\cdot)} :\R^{d}\times \R^{d}\to \R$ is associated 
with a learnable feature map $\Phi:\R^{d}\to \R^{D_\Phi}$.
Here, $\calK(\cdot,\cdot)$
acts column-wise on matrix:  $\calK( \Xi, \bx ) = [ \{\calK( \bxi_\mu, \bx )\}_{\mu=1}^M ] = [  \{\Braket{\Phi(\bxi_\mu),  \Phi(\bx)} \}_{\mu=1}^M]\in\R^M$.
Importantly, KHMs shift the dependency on $\Delta_{\min}$ to $\Delta_{\min}^\Phi$, with
\begin{align*}
    \Delta_{\min}^\Phi \coloneqq \min_{\mu\in[M]} \Delta_\mu^\Phi, \quad \text{where}\quad
    \Delta_\mu^\Phi \coloneqq \calK
    \(\bxi_\mu, \bxi_\mu\) - \underset{\nu,\nu\neq\mu}{\max} \calK\(\bxi_\nu, \bxi_\mu\).
\end{align*}
Notably, $\Delta_{\min}^\Phi$ is learnable and parameterized via $\Phi$.
\citet{wu2024uniform} point out that with $\Phi(\bx) = \bW \bx$, where $\bW \in \R^{d \times D_\Phi}$, finding a suitable $\Phi$ that maximizes $\Delta_{\min}^\Phi$ benefits memory storage.
This construction of $\Phi$ preserves key MHM properties, such as accurate  \citep[Lemma~2.1]{wu2024uniform} and consistent \citep[Thm~2.1]{wu2024uniform} retrieval.
However, direct maximization of  $\Delta_{\min}^\Phi$ is challenging due to its max-min nature.
To circumvent, \citet{wu2024uniform} propose a surrogate loss to maximize $\Delta_{\mu}^\Phi$ \textit{on average} \cite[Def.~2.2]{wu2024uniform}. 
As a result, their approach achieves strong empirical results in memory retrieval for MHMs and supervised learning for transformer models.

Nevertheless, maximizing $\Delta_{\mu}^\Phi$ on average, rather than $\Delta_{\min}^\Phi$, raises questions about how their surrogate loss benefits memory storage.
Moreover, the impact of $\Delta_{\min}^\Phi$ on memory capacity lacks a clear analytical characterization, and no theoretical analysis confirms whether maximizing $\Delta_{\min}^\Phi$ leads to optimal memorization capacity.

In this paper, we address these questions from the perspective of (optimal) spherical codes from information thoery \cite{delsarte1991spherical}.
A spherical code is a set of points (vectors) distributed on the surface of a hypersphere, and an optimal spherical code is when the minimum angular distance between any two points is maximized.
In other words, optimal spherical codes aim to spread the points as evenly as possible over the surface of the sphere.
This aligns with the intuition behind KHM --- increasing average separation between stored memories improves memory capacity.
Therefore, 
we treat the stored memory pattern set as a spherical code (\cref{def:sph-code}), and require this spherical code to satisfy the \textit{well-separation condition} \citep[Thm~3.1]{hu2023SparseHopfield}.
We term this spherical code as \textit{memory code}. 
Surprisingly, 
this unique connection enables a tight analysis on KHMs' capacity.

\textbf{Contributions.} Through the memory code perspective, this work makes three main contributions:
\begin{itemize}
    \item \textbf{Provably Optimal Capacity.}
    We study the optimal memory capacity of KHMs and identify the conditions necessary to achieve it.
    Specifically, we derive a provably tight and optimal capacity by matching the well-known exponential lower bound for the memory capacity of MHMs \cite{wu2024uniform,wu2023stanhop,hu2023SparseHopfield,ramsauer2020hopfield} with an upper bound in the low-temperature region. 
    Notably, we establish this tight bound by showing that KHMs store the most memories when the memory set forms an optimal spherical code (\cref{thm:optimal-code-capacity}).
    This result suggests a tight exponential scaling of memory capacity with the pattern dimension $D_\Phi$ (\cref{prop:capacity-bound}).

    \item \textbf{Fast Algorithm.}
    We introduce an algorithm,  $\mathtt{U}\text{-}\mathtt{Hop}$+, that achieves the optimal capacity of KHM in sublinear time. 
    Theoretically, we show that, as temperature approaches zero, $\mathtt{U}\text{-}\mathtt{Hop}$+ finds the optimal feature map for maximal KHM capacity (\cref{gamma-conv-2}). 
    This result bridges our theoretical findings with practical applications and explains the empirical successes of \cite{wu2024uniform}.

    \item \textbf{Numerical Validation.}
    Our experiments validate our theoretical analysis. 
    We observe that (i) $\mathtt{U}\text{-}\mathtt{Hop}$+ creates distinct low-energy regions for each memory pattern, addressing the memory confusion problem in MHMs \cite{wu2024uniform,krotov2016dense}; (ii) $\mathtt{U}\text{-}\mathtt{Hop}$+  significantly reduces metastable states on both MNIST and synthetic datasets, indicating larger memory capacity; (iii) with $\mathtt{U}\text{-}\mathtt{Hop}$+, the  KHMs update rule  converges to fixed points faster.
    
\end{itemize}

\textbf{Organization.}
\cref{sec:intro} presents a brief review of MHMs and KHMs. \cref{sec:related_work} includes related work discussions.
\cref{sec:method} presents our main results. 
Specifically, \cref{high-prob-bound} presents a memory capacity lower bound for KHMs, \cref{method:spherical-code} presents the optimal capacity bound based on the notation of memory code.
\cref{method:algorithm} presents a sublinear time algorithm to search for the optimal $\Phi$. \cref{method:dimemsion-impact} discusses the relationship between $\Phi$ and $M$. 
\cref{sec:exp} includes numerical experiments.

\textbf{Notations.}
Lower case letters denote (column) vectors and upper case letters denote matrices.
We write $\Braket{\ba, \bb} \coloneqq \ba^\sT \bb$ as the inner product for vectors $\ba, \bb \in \R^d$.
The index set $\{ 1, ..., I \}$ is denoted by $
[I]$, where $I \in \mathbb{N}^+$. 
The spectral norm is denoted by $\norm{\cdot}_2$ which is equivalent to the $\ell_2$-norm when applied to a vector.
We denote the memory patterns (keys) by $\bxi \in \R^d$ and the query pattern by $\bx \in \R^d$, and $\Xi \coloneqq [ \bxi_1, ... ,\bxi_M ] \in \R^{d \times M}$ as shorthand for stored memory patterns $\{ \bxi_\mu \}_{\mu \in [M]}$.
Throughout this work, we use $\Xi$ interchangeably to refer to either a $d \times M$ matrix or a set of $M$ $d$-dimensional memory pattern vectors.

\section{Main Theory}
\label{sec:method}
We provide a theoretical analysis on the optimal memory capacity of KHMs.
First, we begin by comparing the memory capacity between MHM and KHM using the standard high-probability lower bound \cite{hu2023SparseHopfield,ramsauer2020hopfield}. 
Then, we present a spherical code perspective as a framework for depicting the optimal memory capacity of both MHMs and KHMs.
In our analysis, we make the following pattern normalization assumption on memory patterns:\footnote{
It is a common assumption adopted in \cite{santos2024sparse, wu2024uniform}.
A justification is the connection to transformer-attention. 
The modern Hopfield update rule (a.k.a. retrieval dynamics) is constantly compared to attention-mechanism and the input query and memory set correspond to $Q$, $K$ for attention.
As LayerNorm is a commonly used strategy in attention layers, this setup can also be seen in real-world scenarios.
}
\begin{assumption}\label{assumption1}
    We assume memory patterns 
    $\norm{\xi_\mu} = 1$ in the rest of our paper.
\end{assumption}

\subsection{High-Probability Capacity Lower Bound}\label{high-prob-bound}

We start by showing the memory capacity of KHM using the standard capacity lower bound introduced by \citet{ramsauer2020hopfield}.
This provides a direct comparison between KHMs and previous works.
The definition of the generalized fixed point \cite{sriperumbudur2009convergence} is
\begin{definition}[Generalized Fixed Point \cite{sriperumbudur2009convergence}]\label{def:general-fixed-point}
We say a set $S \subseteq \R^d$ is a \textit{generalized fixed point} w.r.t. $\calT$ if $\calT(\by) \in S$ for every $\by \in S$.
\end{definition}
\begin{remark}
    In contrast to \cref{def:general-fixed-point}, a fixed point of $\calT$ is a point $\by$ satisfying $\calT(\by) = \by$.
\end{remark}

Let $S_\mu^\Phi$ be a ball with radius $R_\Phi$\footnote{By definition, neighborhoods do not overlap: $S_\mu^\Phi \cap S_\nu^\Phi = \emptyset$ for $\nu\neq\mu$.} centered at every memory pattern in the feature space $\Phi(\bxi_\mu)$:
\begin{align}\label{eq:min_sep}
    S_\mu^\Phi = \{ \by \;|\; \norm{ \Phi(\bxi_\mu) - \by } \leq R_\Phi \}, 
    \quad\text{where}\quad
    R_\Phi \coloneqq
    \half \underset{\underset{ \mu\neq\nu }{\mu,\nu\in[M]}}{\min} \norm{ \Phi(\bxi_\mu) - \Phi(\bxi_\nu)}.
\end{align}
Following \cite{wu2024uniform}, we define the memory storage and retrieval as:
\begin{definition}[Pattern Storage and Retrieval]
\label{def:stored_and_retrieved}
We say a memory pattern $\bxi_\mu$ is \textit{stored} if $S_\mu^\Phi$ is a generalized fixed point of $\calT$, and there exists a fixed point $\bx^\star_\mu \in S_\mu^\Phi$.
A memory pattern $\bxi_\mu$ gets $\epsilon$\textit{-retrieved} by $\calT$ with an input query $\bx$ if $\norm{\calT(\bx) - \bxi_\mu} \leq \epsilon$.
\end{definition}
This definition is compatible with both KHMs and MHMs (with identity feature map).
Under \cref{def:stored_and_retrieved}, KHM's memory capacity is lower bounded by the following lemma.

\begin{lemma}[Memory Capacity of KHM]
\label{lem:KHM-lower-bound}
    Let $1 - p$ be the probability of successfully storing and retrieving a pattern. Assuming the patterns are normalized, the number of patterns $M_\Phi$ that can be stored and retrieved by the KHM, following the update rule  \eqref{KHM-update-rule}, is lower-bounded by:
    \begin{align*}
        M_\Phi \geq \sqrt{p} C^{(D_\Phi-1)/4},
    \end{align*}
    where $C$ is the solution to $C = \nicefrac{b}{(W_0 (\exp{a + \ln{b}}))}$, with $W_0(\cdot)$ being the principal branch of Lambert $W$ function,
    $a \coloneqq (\nicefrac{4}{(D_\Phi-1)}) \( \ln (\nicefrac{(2\sqrt{p}-2)}{R_\Phi}) + 1\) $ and $b \coloneqq \nicefrac{4\beta}{(5(D_\Phi-1))}$.
    For comparison, $M_\Phi$ reduces to MHM's capacity lower bound by setting $\Phi = I_d$, with $D_\Phi = d$.
\end{lemma}

\begin{proof}\vspace{-.5em}
    Our proof follows \cite{hu2023SparseHopfield, wu2023stanhop}.
    See \cref{proof:capacity-lower-bound} for a  proof.
\end{proof}\vspace{-.5em}

With a fixed $D_\Phi$, the highest lower bound of \cref{lem:KHM-lower-bound} corresponds to specific a $\Phi$ that maximizes $R_\Phi$.
This provides an intuitive insight on the design of separation loss \citep[Definition~2.2]{wu2024uniform} for kernel learning in \citep[Algorithm~1]{wu2024uniform}.
With an additional feature space, KHM has an exponential memory capacity in $D_\Phi$ that does not depend on $d$. 
When $D_\Phi = d$, KHMs obtain a tighter lower bound than MHMs if $R_\Phi > R$. 
This bound connects the storage capacities of KHMs and MHMs, showing that their capacities scale exponentially with respect to $D_\Phi$ and $d$.

\subsection{Memory Code: Memories as Spherical Code}\label{method:spherical-code}

There are two aspects the lower bound in \cref{lem:KHM-lower-bound} does not address: the maximal capacity of KHMs and the flexibility of choosing different $\Phi$ in KHMs. 
Therefore, we present a new framework using spherical codes to take the above perspectives into consideration for further analysis.
We begin by introducing the concepts of spherical code and optimal spherical code.
\begin{definition}[Spherical Code]\label{def:sph-code}
A $d$-dimensional spherical code on the unit sphere $\mathbb{S}^{d-1}$ is a finite set $\calC_N = \{ \bc_1,...,\bc_N\}$ of $\mathbb{S}^{d-1}$ with $N$ points, where 
$c_i\in\R^d$ for $i\in[N]$ and $\abs{\calC_N}=N$.    
\end{definition}

\begin{definition}[Minimal Separation]
The minimal separation $\rho(\calC_N)$ of a spherical code $\calC_N$ is the maximal inner product between two distinct points in $\calC_N$:
\begin{align*}
    \rho(\calC_N) = \underset{ \bc_i,\bc_j \in \calC_N}{\max}
    \Braket{\bc_i, \bc_j}, \quad\text{for every}\;i\neq j.
\end{align*}
\end{definition}

\begin{definition}[Optimal Spherical Code]
Let $\calC_N = \{ \bc_1, \dots, \bc_N \} \subseteq \mathbb{S}^{d-1}$ be a $d$-dimensional spherical code with $N$ points. An optimal spherical code $\calC_N^\star$ minimizes the maximal pairwise inner product, which corresponds to maximizing the minimal separation between points in the code.
Formally, the optimal spherical code $\calC_N^\star$ is defined as:
\begin{align*}
    \calC_N^\star = \underset{ \calC_N \subset \mathbb{S}^{d-1} }{\argmin } \underset{i \neq j}{\max} \Braket{ \bc_i, \bc_j }, \quad \text{for}\; i, j \in [N].
\end{align*}
The minimal separation of the optimal spherical code is denoted as $\rho^\star$.\footnote{The optimal arrangement of most spherical codes is unknown, except for specific pairs of $(d, N)$. A list of known optimal arrangements and minimal separations can be found at \url{http://neilsloane.com/packings/} and in \cite{conway2013sphere}.}
\end{definition}

Next, we recall the function class $\calH$ of the linear feature map introduced by \citet{wu2024uniform}:
\begin{definition}
\label{def:function-class}
    The function class $\calH$ consists of linear maps that satisfy the following properties:
    \begin{enumerate}[leftmargin=2.2em,before=\vspace{-0.2em},after=\vspace{-0.2em}]
        \item For all $\Phi \in \calH$, $\Phi : \mathbb{S}^{d-1} \rightarrow \mathbb{S}^{D_\Phi-1}$ is a linear map defined by a matrix $\bW  \in \R^{d \times D_\Phi}$.
        \item The matrix $\bW$ has full column rank.
        \item When applying $\Phi$ to different inputs:
        \begin{itemize}[before=\vspace{0em},after=\vspace{-0.2em}]
            \item For a vector $\bxi \in \R^d$, $\Phi(\bxi) = \bW^\sT \bxi \in \R^{D_\Phi}$.
            \item For a matrix $\Xi \in \R^{d \times M}$, $\Phi(\Xi) = \left( \Phi(\bxi_1), \dots, \Phi(\bxi_M) \right) \in \R^{ D_\Phi \times M }$.
            \item For a set of vectors $\calV = \{ \bv_1, \dots, \bv_N \}$, $\Phi(\calV) = \{ \Phi(\bv_1), \dots, \Phi(\bv_N) \}$ with $\abs{\Phi(\calV)} = N$.
        \end{itemize}
    \end{enumerate}
\end{definition}

\cref{def:function-class} ensures KHMs with feature map $\Phi
(\cdot)\in\calH$ satisfying the defining characteristics of MHMs: 
accurate  \citep[Lemma~2.1]{wu2024uniform} and consistent \citep[Thm~2.1]{wu2024uniform} retrieval according to \cref{def:stored_and_retrieved}.\footnote{Note that, \cref{def:function-class} is sufficient but not necessary.
It is possible to find a different $\calH'$ such that KHMs with $\Phi\in\calH'$ achieves desiring characteristics as \citep[Lemma~2.1]{wu2024uniform}  and \citep[Thm~2.1]{wu2024uniform}.}
Now, we combine the concept of spherical code and memory storage.
\begin{definition}[Kernelized Well-Separation Condition \cite{wu2024uniform,wu2023stanhop,hu2023SparseHopfield,ramsauer2020hopfield}]\label{def:kernel-sep}
    Given a set of kernelized memory patterns 
    $\Phi(\Xi) = \{ \Phi(\xi_\mu)\}_{\mu=1}^M \subseteq \mathbb{S}^{D_\Phi-1}$, the kernelized memory pattern $\Phi(\xi_\mu)$ satisfies the well-separation condition if the following holds:
    \begin{align}\label{eqn:well-separation}
        \Delta_{\mu}^\Phi \geq \frac{1}{\beta} \ln\left( \frac{2(M-1)}{R_\Phi} \right),
    \end{align}
    where the inverse temperature $\beta$ is given by \eqref{KHM-update-rule}   and $R_\Phi$ is defined by \eqref{eq:min_sep}. 
\end{definition}

The inequality \eqref{eqn:well-separation} is a necessary condition for the $\mu$-th memory to have a well-defined attractor basin.
Hence, the more memories satisfying \eqref{eqn:well-separation} the greater the memory capacity of the model.

\begin{definition}[Memory Code]\label{def:memory-code}
Let $M\in \mathbb{N}_+$, $\beta>0$, $D_\Phi>1$ and $\Phi\in\calH$.
    For any finite set  $\Phi(\Xi) = \{ \Phi(\xi_\mu)\}_{\mu=1}^M \subseteq \mathbb{S}^{D_\Phi-1}$,
    we say the set $\Phi(\Xi)$ is a memory code if all points in $\Phi(\Xi)$ satisfies \eqref{eqn:well-separation}.
    Further, we denote $\Lambda_{D_\Phi}$ as the set of all memory codes in $\mathbb{S}^{D_\Phi - 1}$, including all possible $\Xi$, $\Phi$. 
\end{definition}
Notably, $\Lambda$ includes all the possible pattern sets $\{\Phi(\Xi)\}$ that are able to be stored and retrieved by kernelized Hopfield models and modern Hopfield models.
Naturally, the optimal memory capacity is the size of the largest memory code in $\mathbb{S}^{D_\Phi-1}$.
This  leads to our next definition:
\begin{definition}[Optimal Memory Capacity]\label{lemma:optimal}
For $D_{\Phi}>1$ and $\Phi\in\calH$,
the optimal capacity $M^\star$ is the  cardinality of the largest memory code in $\Lambda_{D_{\Phi}}$, i.e.,  $M^\star \coloneqq \max_{\Phi(\Xi) \in \Lambda} \abs{ \Phi(\Xi)}$ for all possible $\Xi$,  $\Phi$.
\end{definition}
\cref{lemma:optimal} specifies the largest possible memory code in $\Lambda_{D_\Phi}$ for a given $D_\Phi$. 
Let $\Tilde{\Xi}$ denote the memory set associated with $M^\star$, such that $\|\Tilde{\Xi}\| = M^\star$. 
To store all patterns in $\Tilde{\Xi}$, we need to find a suitable feature map $\Tilde{\Phi}$ such that $\Tilde{\Phi}(\Tilde{\Xi})$ is a valid memory code.

Following this definition, we present the next lemma and proposition on optimal memory capacity.
\begin{lemma}[Capacity of Optimal Spherical Code]\label{thm:optimal-code-capacity}
    Given a fixed $D_\Phi > 1$, and its corresponding $M^\star$,
    if an optimal code $\calC_{\text{opt}}$ is in $\mathbb{S}^{D_\Phi-1}$ and has size $M^\star$,
    then $\calC_{\text{opt}} \in \Lambda_{D_\Phi}$.
\end{lemma}
\begin{proposition}[Optimal Memory Capacity]\label{prop:capacity-bound}
    Following \cref{thm:optimal-code-capacity},  we have
    \begin{align*}
    M^\star \asymp c^{D_\Phi},     
    \end{align*}
    for some $c>1$. 
    Here $\asymp$ indicates matching upper and lower bounds up to constant factors.
\end{proposition}
\begin{proof}[Proof Sketch]
    We proof \cref{thm:optimal-code-capacity} by showing that the model capacity is a increasing function w.r.t. the minimal separation value.
    For \cref{prop:capacity-bound}, we utilize the upper bound in \cite{kabatiansky1978bounds} and lower bound in \cite{wyner1965capabilities, shannon1959probability, chabauty1953resultats} to bound the quantity.
    Please see \cref{proof-of-optimal-code-capacity} for a  detailed proof.
\end{proof}\vspace{-.5em}
\cref{prop:capacity-bound} indicates that the optimal capacity of MHMs and KHMs scales exponentially with $D_\Phi$.
This capacity bound is provably tight and optimal for large feature dimension $D_\Phi$.
It echos the exponential capacity lower bound in \cref{lem:KHM-lower-bound} and in prior works \cite{wu2024uniform,wu2023stanhop,hu2024outlier,hu2024nonparametric,hu2024computational,hu2023SparseHopfield,ramsauer2020hopfield}.
Moreover,
\cref{thm:optimal-code-capacity} shows that achieving the maximal capacity in any $D_\Phi$ is equivalent to achieving optimal codes.
Thus, for a given memory set $\Xi$ of size $M$, the memory storage problem with KHMs divides into two sub-problems:
\begin{itemize}[leftmargin=2.4em]
    \item [\textbf{(P1)}] \label{item:P1}
    Finding a sufficiently large $D_\Phi$ (in \cref{method:dimemsion-impact}), and

    \item [\textbf{(P2)}] \label{item:P2}
    Finding a $\Phi$ such that $\Phi( \Xi)$ is an optimal spherical code (in \cref{method:algorithm}).

\end{itemize}

Next, we examine these two sub-problems and present a sub-linear time algorithm to solve them.

\section{Sub-Linear Time Algorithm for Optimal Memory Capacity}
\label{sec:algo}
In this section, we  present an sub-linear time algorithm that achieves optimal capacity.
Then, we analyze the scaling behavior of $D_\Phi$ for KHMs to store any desired amount of memories.

\subsection{Learning to Achieve Optimal Memory Code}\label{method:algorithm}

Here we present an asymptotic result showing that an algorithm exists to find the optimal $\Phi$ for maximizing memory storage in dimension $D_\Phi$.
Building on the results from the previous sections, we consider the following problem:
\begin{problem}[HardMax Problem]\label{hardmax}
    Given a memory set $\Xi = \{ \bxi_1, \dots, \bxi_M \}$,
    and assuming that $D_\Phi$ is sufficiently large to satisfy \eqref{eqn:well-separation}, we define the HardMax problem  as finding a $\Phi$ such that $\Phi(\Xi)$ forms an optimal spherical code:
    \begin{align}\label{eqn:hardmax}
    \min_{\Phi \in \calH}
    \calL_{\text{HardMax}}(\Phi), \quad \text{where} \quad
    \calL_{\text{HardMax}}(\Phi) \coloneqq
    \max_{\nu, \mu \in [M], \nu \neq \mu}
      \left\langle \Phi(\xi_\mu), \Phi(\xi_\nu) \right\rangle \geq \rho^\star.
    \end{align}
\end{problem}

This problem setup involves finding a $\Phi$ such that $\Phi(\Xi)$ forms an optimal spherical code. 
Ideally, a more expressive function class $\calH$ would simplify finding such a $\Phi$; exploring explicit forms of more powerful mappings is left for future work. 
Note that \eqref{eqn:hardmax} represents a min-max optimization problem. 
Achieving the global optimum is notoriously challenging \cite{hsieh2021limits, daskalakis2021complexity, shen2020learning}.
Thus, we introduce a surrogate objective to solve \eqref{eqn:hardmax}:
\begin{definition}[Average Separation Loss]\label{def:cont-sep-loss}
    For $\tau > 0$, given a set of memory patterns $\Xi$ and a feature map $\Phi$, we define the average separation loss as
    \begin{equation}\label{eqn:sep-loss}
    \calL( \Xi, \tau, \Phi ) \coloneqq
        \frac{1}{M} \sum_{\mu=1}^M \ell_\mu( \Xi, \Phi, \tau ),
        \;
        \text{where}
        \;
        \ell_\mu(  \Xi, \Phi, \tau ) \coloneqq
        \log \[
         \sum_{\nu=1}^M \exp( \frac{\langle \Phi(\xi_\mu), \Phi(\xi_\nu) \rangle}{\tau}  )  
        \].
    \end{equation}
\end{definition}
The primary difference between \eqref{eqn:hardmax} and \eqref{eqn:sep-loss} is that  \eqref{eqn:sep-loss} calculates average separation, whereas \eqref{eqn:hardmax} focuses on the maximum separation between a single pair.
This surrogate loss alleviates the challenging optimization, as \eqref{eqn:sep-loss} is convex.
Therefore, with vanishing temperature $\tau$, the next theorem shows that \eqref{eqn:sep-loss} converges to the HardMax problem asymptotically.
\begin{theorem}
\label{gamma-conv-2}
    For any possible integer $M$, we have 
    \begin{align*}
        \underset{\tau\rightarrow0}{\lim\sup}   
        \(
        \argmin_{ \Phi \in \calH }
        \calL(\Xi, \Phi, \tau )
         \) \subseteq
        \argmin_{ \Phi \in \calH } \calL_{\text{HardMax}}(\Phi).     
    \end{align*}
\end{theorem}
\begin{proof}\vspace{-.5em}
    we first introduce a helper function $\calL_0$ in \eqref{eqn:helper}. 
    We show that as $\tau \rightarrow 0$, $\calL_0$ converges uniformly to $\calL_{\text{HardMax}}$. Then, we prove that optimizing $\calL_0$ and $\calL$ yields the same optimal solution.
    Please see \cref{proof:gamma-conv-2} for a detailed proof.
    \vspace{-.5em}
\end{proof}
\cref{gamma-conv-2} indicates that, with vanishing temperature, the minimiozation of \eqref{eqn:sep-loss} converges to the HardMax problem, i.e., their share the same optimal solution.
This provide a theoretical justification for the empirical success of \cite{wu2024uniform}.
In particular, the surrogate objective -- the maximizing average separation between memories --- leads to provably optimal memory capacity in low-temperature region (i.e., $\tau \to 0$).
Lastly, we remark that that this analysis provides theoretical insights rather than practical guidance.
To achieve high retrieval accuracy, the setting ($\tau=1$) in \cite{wu2024uniform} is sufficient for a wide range of applications.

\paragraph{$\mathtt{U}\text{-}\mathtt{Hop}$+: Sub-Linear Time Algorithm for Achieving Optimal Memory Capacity.} 
Next, we present \cref{algorithm1} for finding a $\Phi$ such that $\Phi(\Xi)$ forms an optimal spherical code. 
To meet the conditions in \cref{def:function-class}, we use projected gradient descent to convert this constrained optimization problem into an unconstrained one.
Several methods satisfy the requirements in \cref{def:function-class}; we discuss them in \cref{pgd}. 
We denote the learning rate as $\gamma$, the input matrix of the loss function as $\bX$, and the weight matrix as $\bW$. 
We define a single \underline{P}rojected \underline{G}radient \underline{D}escent (PGD) step as
\begin{align*}
    \bW_{t+1} = \mathtt{PGD}( \bW_t, \gamma, \bX),
\end{align*}
We defer the detailed formulation to \cref{pgd}.
\begin{algorithm}[!t]
\caption{$\mathtt{U}\text{-}\mathtt{Hop}$+}\label{algorithm1}
\textbf{Input:}  Iterations $N$, feature map $\Phi(\bx) \coloneqq \bW \bx$, memory set $\Xi$, learning rate $\gamma \leq 1/G$ where $G$ is the Lipschitz constant of $\calL$  \\
\textbf{Output:} $\bx$  
\begin{algorithmic}[1]
\STATE $\bW_0 \leftarrow \bW$
\FOR{$t = 0, ... N-1$}  
    \STATE $\bW_{t+1} \leftarrow \mathtt{PGD}
    \( \bW_t, \gamma, \Xi \).$
\ENDFOR
\STATE return $\bW_N$
\end{algorithmic}
\end{algorithm}
Since the separation loss is convex and smooth, using projected gradient descent with a learning rate $\gamma \leq 1/G$, yields a sub-linear convergence rate of $\calO(\nicefrac{1}{N})$ \cite{iusem2003convergence}.
This provides an asymptotic solution to the first sub-problem (\hyperref[item:P1]{(P1)}).
Next, we examine the relationship between feature dimension $D_\Phi$ and the number of memories $M$.

\subsection{Impact of \texorpdfstring{$D_\Phi$}{}}\label{method:dimemsion-impact}

This subsection analyzes the minimum $D_\Phi$ to store a given set of $M$ memories.
Based on the \textit{well-separation condition} and the derivation in \cref{proof-of-optimal-code-capacity}, the required $\Delta_{\min}^\Phi$ to store $M$ memories scales as $\calO( \ln(M) )$.
With this insight, the following proposition shows the scaling behavior of required $D_\Phi$ with respect to $M$ and $\Delta_{\min}^\Phi$.
\begin{proposition}\label{lemma:separation-bound}
    Let $M^\star$ be the optimal memory capacity in $\mathbb{S}^{D_\Phi}$ and $\Phi\in\calH$.
    For any optimal code $\bC^\star$ in $\mathbb{S}^{D_\Phi-1}$ of size $M^\star$, the minimal separation $\rho(\bC^\star)$ is bounded by:
    \begin{align*}
        \frac{1}{2}
        \left( 
        \frac{\sqrt{\pi}}{M^\star}
        \cdot 
        \frac{\Gamma\left( \frac{D_\Phi+1}{2} \right)}{ \Gamma\left( \frac{D_\Phi}{2} + 1 \right) }
        \right)^{\frac{2}{D_\Phi-1}}
        \leq
        \underset{\Phi \in \calH}{\max}\;
        \Delta_{\min}^\Phi
        \leq
        2
        \left(
        \frac{2\sqrt{\pi}}{M^\star}
        \cdot \frac{ \Gamma \left( \frac{D_\Phi+1}{2} \right) }{ \Gamma \left( \frac{D_\Phi}{2} \right) }
        \right)^{\frac{1}{D_\Phi-1}},
    \end{align*}
    where $\Gamma(\cdot)$ is the gamma function.
\end{proposition}
\begin{remark}
    By  gamma function asymptotics, 
    \cref{lemma:separation-bound} is consistent with 
    \cref{prop:capacity-bound}.
\end{remark}

\begin{proof}\vspace{-.5em}
    Please see \cref{proof:separation-bound} for a detailed proof.
\end{proof}
\vspace{-.5em}
\cref{lemma:separation-bound} establishes the separation value for memory codes that achieve optimal capacity in $D_\Phi$-dimensional space.
Using this bound, for a given separation value $\Delta_{\min}^\Phi$, the minimum $D_\Phi$ required to store $M$ points scales as $\log (M^2/\Delta_{\min}^\Phi)$.
We conduct an experiment to demonstrate the bound’s tightness and provide an example with
 $D_\Phi=3$ in \cref{fig:sep-bound}.

\section{Experimental Studies}
\label{sec:exp}

\begin{table*}[t]
\vspace{-2em}
    \centering
    \setlength{\cmidrulekern}{0.25em}
    \caption{\small\textbf{Distribution of Metastable State (in $\%$).} 
    For MNIST, we use the training set as memories and test set as queries.
    For synthetic data, we randomly generate the memories and queries.
    $\norm{\bp}_0$ denotes the size of metastable state , which is the amount of non-zero entries of the probability distribution. 
    For Softmax, we use a threshold of $0.01$. For hyperparameter settings, see \cref{table:hyper-meta}.
    }
    \resizebox{ \textwidth}{!}{%
    \begin{tabular}{ccccccccccccc}
    \toprule 
         & \multicolumn{6}{c}{Synthetic} &  \multicolumn{6}{c}{MNIST}
        \\
        \cmidrule(lr){2-7}                  
        \cmidrule(lr){8-13}
        & \multicolumn{2}{c}{Softmax}
        & \multicolumn{2}{c}{$1.5$-entmax}
        & \multicolumn{2}{c}{sparsemax}
        & \multicolumn{2}{c}{Softmax}
        & \multicolumn{2}{c}{$1.5$-entmax}
        & \multicolumn{2}{c}{sparsemax}
        \\
        \cmidrule(lr){2-7}                  
        \cmidrule(lr){8-13}
        $\norm{\bp}_0$        
        & -  & $\mathtt{U}\text{-}\mathtt{Hop}$+
        & -  & $\mathtt{U}\text{-}\mathtt{Hop}$+
        & -  & $\mathtt{U}\text{-}\mathtt{Hop}$+
        & -  & $\mathtt{U}\text{-}\mathtt{Hop}$+ 
        & -  & $\mathtt{U}\text{-}\mathtt{Hop}$+
        & -  & $\mathtt{U}\text{-}\mathtt{Hop}$+
        \\
        \hline
        1 & $0.0$ & \cellcolor{LightCyan} $90.0$ & $0.0$ & \cellcolor{LightCyan} $100.0$ &$0.0$  & \cellcolor{LightCyan}$100.0$ &  $3.48$ &\cellcolor{LightCyan} $100.0$ & $69.2$ & \cellcolor{LightCyan} $100.0$ & $88.1$ & \cellcolor{LightCyan} $100.0$ \\
        2 & $0.0$ & $8.0$ & $0.0$ & $0.0$& $20.0$  &  $0.0$ & $2.16$ & $0.0$  & $8.6$& $0.0$  & $5.2$& $0.0$ \\
        3 & $0.0$ & $0.0$ & $0.0$&$0.0$& $30.0$  &$0.0$ & $1.57$ & $0.0$ & $3.9$ & $0.0$  & $2.6$ & $0.0$\\
        4 & $0.0$ & $2.0$ & $0.0$&$0.0$& $50.0$  &$0.0$ & $1.23$ & $0.0$ & $2.3$& $0.0$  & $1.6$ & $0.0$\\
        5 & $6.0$ & $0.0$ & $0.0$&$0.0$&$0.0$  &$0.0$ & $1.2$ & $0.0$ & $1.6$ & $0.0$  & $1.1$ & $0.0$\\
        6 & $10.0$ & $0.0$& $25.0$&$0.0$&$0.0$ & $0.0$ & $0.95$ & $0.0$ & $0.9$ & $0.0$  & $0.8$ & $0.0$\\
        7 & $20.0$ &$0.0$ & $25.0$&$0.0$& $0.0$ &$0.0$ & $1.04$ & $0.0$ & $0.6$& $0.0$  & $0.4$ & $0.0$\\
        8 & $16.0$ &$0.0$ & $0.0$&$0.0$&$0.0$  &$0.0$ & $0.84$ & $0.0$ & $0.6$ & $0.0$  & $0.1$& $0.0$ \\
        9 & $20.0$ &$0.0$ & $22.5$&$0.0$&$0.0$  &$0.0$ & $1.03$ & $0.0$ & $0.3$& $0.0$  & $0.0$ & $0.0$\\
        10$^+$ & $28.0$ & $0.0$ & $27.5$ &$0.0$&$0.0$ &$0.0$ & $86.5$ & $0.0$ & $12.0$ & $0.0$ & $0.1$& $0.0$ \\
        \bottomrule
    \end{tabular}
    }
    \label{tab:meta-dist}
    \vspace{-1.5em}
\end{table*}

\subsection{\texorpdfstring{$\mathtt{U}\text{-}\mathtt{Hop}$}{}+ Reduces Metastable States}

We compare the distribution of metastable state size under standard MHM and KHM update rules.
The results are in \cref{tab:meta-dist}.
In general, with more metastable state having the size of $1$, meaning the Hopfield model stores more memories as the query converges to a single memory. 
For metastable state size larger than 1, it represents that the retrieved pattern converges near the mean of a subset of memories, violating the requirement of $S_\mu^\Phi \cap S_\mu^\Phi = \emptyset$.

\textbf{Baselines.}
We compare different variants of MHMs and KHMs. \citet{santos2024sparse, wu2023stanhop} provide comprehensive analyses of modern Hopfield models with various normalization functions. 
Here, we consider softmax, 1.5-entmax, and sparsemax for normalization. 
We equip these three baselines with $\mathtt{U}\text{-}\mathtt{Hop}$ to compare against standard MHMs.

\textbf{Settings and Metrics.}
Let $\bp = \Softmax(\beta x^\sT \Xi)$. 
We determine whether the update rule converges to either a single memory or a mixture of memories by observing the probability distribution $\bp$.
The quantity $\norm{\bp}_0$ represents the size of the metastable state, which is the number of non-zero entries in the probability distribution. 
In the case of $1.5$-entmax and sparsemax, we calculate $\norm{\bp}_0$ directly. 
For softmax, since it only generates non-zero entries, we use a threshold of $0.01$ and consider the entries under the threshold as $0$.
We conduct experiments using both synthetic and MNIST datasets. 
For MNIST, we use the training set as memories and the test set as queries. 
For synthetic datasets, we randomly generate memories and queries with Gaussian initialization.
To ensure the convergence to the fixed point, we perform multiple updates on the query. 
For more details, refer to \cref{appendix:meta-detail}.

\textbf{Results.}
On both synthetic and MNIST datasets, it is evident that under separation maximization, the size of the metastable state dramatically decreases within just 20 iterations of \cref{algorithm1}. 
This result demonstrates that, with \cref{algorithm1}, KHMs are capable of storing patterns that MHMs cannot store. 
The significant percentage of size 1 metastable states in KHMs indicates that they circumvent the memory confusion problem in dense associative memory models \cite{krotov2016dense}.
For the MNIST dataset, we see MHMs show close performance with KHMs under $1.5$-entmax and sparsemax, showing that the methods in \cite{santos2024sparse,wu2023stanhop, hu2023SparseHopfield} also circumvent the memory confusion problem.
Notably, KHMs require only one-fourth of the dimensions to store memories while perfectly storing 60,000 MNIST patterns. 
These results suggest that KHMs with \cref{algorithm1} efficiently utilize feature dimensions for memory storage.

\subsection{Energy Landscape under \texorpdfstring{$\mathtt{U}\text{-}\mathtt{Hop}$}{}+ Stores More Memories}

\begin{figure*}[!t]
   \vspace{-5em}
    \centering
    \includegraphics[width=\textwidth]{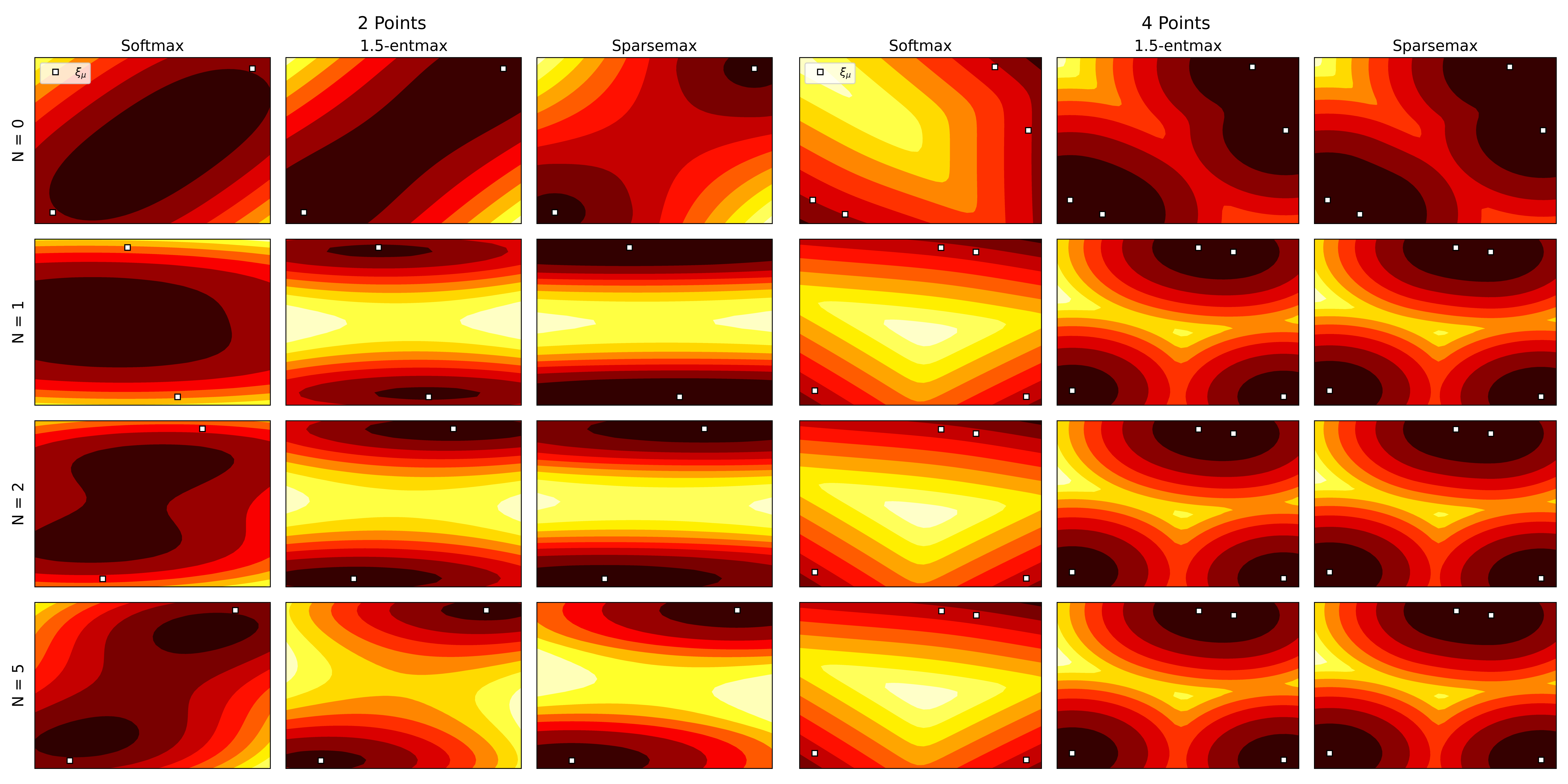}
    \vspace{-1em}
    \caption{\small
    \textbf{Energy Landscape under Different Iterations of \cref{algorithm1}.}    
    Left: $M=2$, Right: $M=4$.
    Lighter color represents higher energy.
    The first row represents the raw energy landscape without applying $\mathtt{U}\text{-}\mathtt{Hop}$+.
    The second to last row represents the energy landscape when $N = (1, 2, 5)$.
    The visualization shows that \cref{algorithm1} not only separates the local minima better, but also pushes memories closer to the fixed point.    }
    \label{fig:contours}
    \vspace{-0.5em}
\end{figure*}

\textbf{Settings and Metrics.}
We visualize the energy landscape of KHMs at different stages of \cref{algorithm1} using contour plots. 
The results are presented in \cref{fig:contours}. We consider two settings: 2 and 4 memories stored in a 2-dimensional space.
Ideally, the energy landscape should position memories in multiple separated low-energy regions (valley), with each region isolated from others by high-energy regions. 
If multiple memories share the same valley, it  leads to memory confusion and the presence of metastable states during energy minimization.
For experiment details, refer to \cref{appendix:landscape-detail}.

\begin{figure*}[t]
    \centering
    \includegraphics[width=0.97\textwidth]{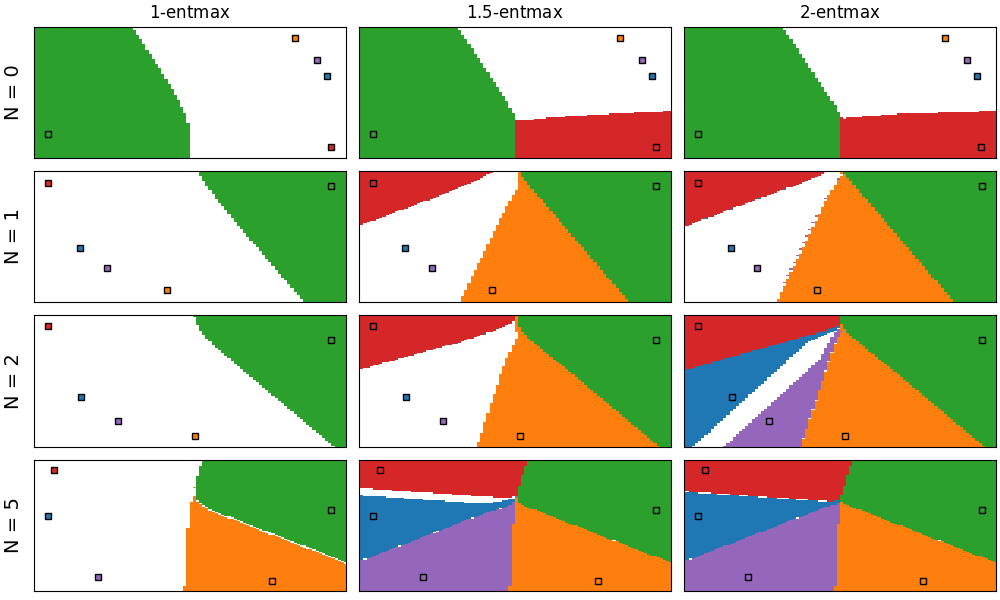}
    \caption{\small
    \textbf{Basins of Attraction Comparison of \cref{algorithm1}.}    
    The first row represents the raw Basins of Attraction without applying $\mathtt{U}\text{-}\mathtt{Hop}$+ or KHM.
    The second to last row shows the basins when $N = (1, 2, 5)$.
    Square points are memories.
    White area is where queries are not able to converge to a single memory.
        Colored area is where queries converges to the corresponding memory.
    The result indicates that $\mathtt{U}\text{-}\mathtt{Hop}$+ is capable of converging to fixed point fast and reduce metastable states. $1$ and $2$-entmax corresponds to Softmax \cite{ramsauer2020hopfield} and Sparsemax \cite{hu2023SparseHopfield}.
    }
    \label{fig:basins}
    \vspace{-1.5em}
\end{figure*}

\textbf{Results.}
The first row in \cref{fig:contours} shows the raw energy landscape without KHM and \cref{algorithm1}, corresponding to the modern Hopfield energy landscape. 
On the right side of \cref{fig:contours}, we observe that MHMs are only able to store 2 out of 4 points, but KHMs are able to further separate one point from the others, resulting in memorizing one extra pattern. 
\cref{thm:optimal-code-capacity} and \cref{gamma-conv-2} indicate that $\mathtt{U}\text{-}\mathtt{Hop}$+  pushes memories away from each other, providing more isolated $S_\mu^\Phi$, for $\mu\in[M]$. 
We observe this phenomenon across all settings, especially under the 2-point configuration with Softmax and $1.5$-entmax, where the low-energy region is split into two distinct valleys as $N$ increases. 
This process shows how $\mathtt{U}\text{-}\mathtt{Hop}^+$ is able to store memories that MHMs cannot.
With energy minimization, the query converges to either one of the minima instead of the mixture thereof (also showed in \cref{fig:basins}).
Additionally, we also notice that the contour lines exhibit steep slopes between different local minima in the 2-point setting under $1.5$-entmax and sparsemax. 
This implies that $\mathtt{U}\text{-}\mathtt{Hop}$+ pushes local minima further away from each other and deepens each one of them. 
Such sharp changes in energies lead to faster convergence to fixed points due to larger gradients.

\textbf{Basins of Attraction.}
\cref{fig:basins} shows the basins of attraction of queries w.r.t. MHM and KHM under the scenario of storing 5 patterns.
We randomly initialize 5 patterns with normal distribution.
We run the update rule for $5$ iterations and see whether each query converges to a single memory (colored) or to a metastable state (white).
Following the above setting, we track the attraction basins throughout each iteration of \cref{algorithm1}.
We defer more details to \cref{appendix:basins}.
Specifically, most MHM variants are not capable of converging to fixed points in $5$ updates.
While $\mathtt{U}\text{-}\mathtt{Hop}$+ dramatically improves such aspect, where most queries are able to converge either one of the memories.
Moreover, the increased $R_\Phi$ also leads to a larger $S_\mu^\Phi$, making more queries to converge to a single memory.
Additionally, there is a performance gap between Softmax ($\alpha=1$) and other sparse variants, which matches the findings in \cite{santos2024sparse, hu2023SparseHopfield}.

\begin{table}[t]
    \centering
    \setlength{\cmidrulekern}{0.25em}
    \caption{\small{
    \textbf{Test AUC of Multiple Instance Learning Datasets.} 
    We compare the $\mathtt{HopfieldPooling}$-based model with and without \uhop+.
    We use the dense \cite{ramsauer2020hopfield} and sparse \cite{hu2023SparseHopfield} modern Hopfield models as baselines.
    We use $K$-fold cross validation on all 4 datasets, with $K = 10$.
    The reported AUC is the average AUC score across 10 folds.
    For the baselines, we use the results reported in \cite{hu2023SparseHopfield}.
    For our method, we directly use the default hyperparameter without grid search instead of using hyperparameter optimization (HPO) in \cite{hu2023SparseHopfield, ramsauer2020hopfield}.
    We exclude the variance as they are all smaller than $0.07$.
    The result shows that even without HPO, \uhop+ is still able to obtain a performance gain.
    }
    }
    \begin{tabular}{lcccc}
    \toprule 
         \small{Method}  &  \small{Tiger} & \small{Elephant} & \small{Fox} & \small{UCSB} \\
         \hline
         \small{Modern Hopfield} & $0.871$ & $0.876$ & $0.637$ &  $0.828$ \\
         \small{Modern Hopfield + \uhop+} & \cellcolor{LightCyan}$0.881$ & \cellcolor{LightCyan}$0.921$ & \cellcolor{LightCyan}$0.648$ & \cellcolor{LightCyan}$0.831$ \\
         \midrule
        \small{Sparse Hopfield} & $0.884$ & $0.914$ & $0.610$ &  $0.796$ \\
         \small{Sparse Hopfield + \uhop+} & \cellcolor{LightCyan}$0.887$ & \cellcolor{LightCyan}$0.921$ & \cellcolor{LightCyan}$0.638$ & \cellcolor{LightCyan}$0.805$ \\
         \bottomrule
    \end{tabular}
    \label{tab:mil}
    \vspace{-1em}  %
\end{table}

\subsection{Multiple Instance Learning}
We conduct multiple instance learning (MIL) on 4 real-world datasets using Hopfied-based models with and without our \uhop+ algorithm.
We follow the setup in \cite{santos2024sparse, wu2024uniform, hu2023SparseHopfield} by using a model with 1 embedding layer, 1 $\mathtt{HopfieldPooling}$ layer and a linear readout layer.
We first utilize \cref{algorithm1} to ``pretrain'' the embedding and $\mathtt{HopfieldPooling}$ layer, and then fine-tune the whole model on the MIL task.
The results are in \cref{tab:mil}.
We observe that both dense and sparse Hopfield-based models obtain performance boost when equipped with \uhop+, indicating our method is also effective in practical scenarios.
Further, as demonstrated in \cref{fig:loss-conv}, the separation loss converges fast, indicating \uhop+ is a lightweight method for performance boost.

\section{Discussion and Conclusion}
\label{sec:conclusion}

This work complements $\mathtt{U}\text{-}\mathtt{Hop}$ \cite{wu2024uniform} by 
establishing the optimal capacity of kernelized modern Hopfield models (KHMs) and providing the first tight, optimal memory capacity bound for transformer-compatible dense associative memories.
We start by connecting stored memories in KHMs to spherical codes from information theory.
We then prove that maximizing memory storage in KHMs requires arranging memories as an optimal spherical code in feature space.
This allows us to matches the well-known  exponential lower bound \cite{wu2024uniform,wu2023stanhop,hu2024outlier,hu2024nonparametric,hu2024computational,hu2023SparseHopfield,ramsauer2020hopfield} with an upper bound.
This achievement is notable, as deriving such a tight bound is challenging due to the max-min structure of maximal separation among stored memories \cite[Section~5]{wu2024uniform}.
Moreover, we introduce a sub-linear time algorithm to achieve this optimal capacity, $\mathtt{U}\text{-}\mathtt{Hop}$+ (\cref{algorithm1}).
$\mathtt{U}\text{-}\mathtt{Hop}$+ performs this rearrangement with a convergence rate of $\mathcal{O}(\nicefrac{1}{N})$.
Additionally, we analyze the minimum dimension $D_\Phi$ required to store $M$ memories. 
Numerically, we validate the effectiveness of KHMs and demonstrate how \cref{algorithm1} enhances memory storage in both KHM retrieval tasks and transformer representation learning tasks.

\textbf{Can $\mathtt{U}\text{-}\mathtt{Hop}$+ Preserve Semantic Meanings?}
In representation learning, it is crucial to preserve relationships in the feature space after encoding data \cite{wang2023improving,neelakantan2022text}. 
The primary strategy is to ensure the embeddings of similar instances share similar directions in Euclidean space. 
At first glance, the approach of pushing all memories away from each other in \cref{eqn:sep-loss} may seem counterintuitive. 
However, as detailed in \cref{appendix:assignment}, we find that the learned feature map still encodes similar instances closely together (\cref{fig:assignment}), even without semantic information involved. 
This result indicates that \uhop+ stores memories in a semantically coherent manner.
A discussion of the separation capability of  $\Phi$ can be found in \cref{fig:assignment}.

\textbf{Limitations.}
One limitation of our work only considers linear affine functions as the feature map $\Phi\in\calH$.
Additionally, standard spherical code analysis focuses only on normalized points on a hypersphere, ignoring memories with varying magnitudes. 
We leave them for future research.

\clearpage

\section*{Broader Impact}
We expect no negative social impacts as this work mostly present theoretical results and numerical simulations.
As discussed in our introduction, this paper develops a theoretical framework to study Kernelized Hopfield models, potentially benefit the area of computational associative (Hopfield) memory models, transformer networks and large foundation models.

\section*{Acknowledgements}
JH thanks Thomas Burns, Dmitry Krotov, Dino Feng, and Andrew Chen for enlightening discussions; Robin Luo, Jiahao Yu, Weimin Wu, and Teng-Yun Hsiao for collaboration on related topics; the Red Maple Family for their support; and Jiayi Wang for facilitating experimental deployments. The authors also thank the anonymous reviewers and program chairs for their constructive comments.

JH is partially supported by the Walter P. Murphy Fellowship.
DW is supported by NIH R01LM1372201.
HL is partially supported by NIH R01LM1372201, AbbVie and Dolby.
This research was supported in part through the computational resources and staff contributions provided for the Quest high performance computing facility at Northwestern University which is jointly supported by the Office of the Provost, the Office for Research, and Northwestern University Information Technology.
The content is solely the responsibility of the authors and does not necessarily represent the official
views of the funding agencies.

\def\arxivfont{\rm}
\bibliographystyle{plainnat}

\bibliography{refs}

\begin{thebibliography}{91}
\providecommand{\natexlab}[1]{#1}
\providecommand{\url}[1]{\texttt{#1}}
\expandafter\ifx\csname urlstyle\endcsname\relax
  \providecommand{\doi}[1]{doi: #1}\else
  \providecommand{\doi}{doi: \begingroup \urlstyle{rm}\Url}\fi

\bibitem[Achiam et~al.(2023)Achiam, Adler, Agarwal, Ahmad, Akkaya, Aleman, Almeida, Altenschmidt, Altman, Anadkat, et~al.]{achiam2023gpt}
Josh Achiam, Steven Adler, Sandhini Agarwal, Lama Ahmad, Ilge Akkaya, Florencia~Leoni Aleman, Diogo Almeida, Janko Altenschmidt, Sam Altman, Shyamal Anadkat, et~al.
\newblock Gpt-4 technical report.
\newblock \emph{arXiv preprint arXiv:2303.08774}, 2023.

\bibitem[Auer et~al.(2023)Auer, Gauch, Klotz, and Hochreiter]{auer2023conformal}
Andreas Auer, Martin Gauch, Daniel Klotz, and Sepp Hochreiter.
\newblock Conformal prediction for time series with modern hopfield networks.
\newblock \emph{Advances in Neural Information Processing Systems}, 36:\penalty0 56027--56074, 2023.

\bibitem[Bai et~al.(2023)Bai, Chen, Wang, Xiong, and Mei]{bai2023transformers}
Yu~Bai, Fan Chen, Huan Wang, Caiming Xiong, and Song Mei.
\newblock Transformers as statisticians: Provable in-context learning with in-context algorithm selection.
\newblock \emph{arXiv preprint arXiv:2306.04637}, 2023.

\bibitem[Bietti et~al.(2024)Bietti, Cabannes, Bouchacourt, Jegou, and Bottou]{bietti2024birth}
Alberto Bietti, Vivien Cabannes, Diane Bouchacourt, Herve Jegou, and Leon Bottou.
\newblock Birth of a transformer: A memory viewpoint.
\newblock \emph{Advances in Neural Information Processing Systems}, 36, 2024.

\bibitem[Bommasani et~al.(2021)Bommasani, Hudson, Adeli, Altman, Arora, von Arx, Bernstein, Bohg, Bosselut, Brunskill, et~al.]{bommasani2021opportunities}
Rishi Bommasani, Drew~A Hudson, Ehsan Adeli, Russ Altman, Simran Arora, Sydney von Arx, Michael~S Bernstein, Jeannette Bohg, Antoine Bosselut, Emma Brunskill, et~al.
\newblock On the opportunities and risks of foundation models.
\newblock \emph{arXiv preprint arXiv:2108.07258}, 2021.

\bibitem[Braides(2006)]{braides2006handbook}
Andrea Braides.
\newblock A handbook of $\gamma$-convergence.
\newblock In \emph{Handbook of Differential Equations: stationary partial differential equations}, volume~3, pages 101--213. Elsevier, 2006.

\bibitem[Bubeck et~al.(2023)Bubeck, Chandrasekaran, Eldan, Gehrke, Horvitz, Kamar, Lee, Lee, Li, Lundberg, et~al.]{bubeck2023sparks}
S{\'e}bastien Bubeck, Varun Chandrasekaran, Ronen Eldan, Johannes Gehrke, Eric Horvitz, Ece Kamar, Peter Lee, Yin~Tat Lee, Yuanzhi Li, Scott Lundberg, et~al.
\newblock Sparks of artificial general intelligence: Early experiments with gpt-4.
\newblock \emph{arXiv preprint arXiv:2303.12712}, 2023.

\bibitem[Burns(2024)]{burns2024semantically}
Thomas~F Burns.
\newblock Semantically-correlated memories in a dense associative model.
\newblock \emph{arXiv preprint arXiv:2404.07123}, 2024.

\bibitem[Burns and Fukai(2023)]{burns2023simplicial}
Thomas~F Burns and Tomoki Fukai.
\newblock Simplicial hopfield networks.
\newblock \emph{arXiv preprint arXiv:2305.05179}, 2023.

\bibitem[Cabannes et~al.(2023)Cabannes, Dohmatob, and Bietti]{cabannes2023scaling}
Vivien Cabannes, Elvis Dohmatob, and Alberto Bietti.
\newblock Scaling laws for associative memories.
\newblock \emph{arXiv preprint arXiv:2310.02984}, 2023.

\bibitem[Chabauty(1953)]{chabauty1953resultats}
Claude Chabauty.
\newblock Resultats sur lempilement de calottes egales sur une perisphere de rn et correction a un travail anterieur.
\newblock \emph{COMPTES RENDUS HEBDOMADAIRES DES SEANCES DE L ACADEMIE DES SCIENCES}, 236\penalty0 (15):\penalty0 1462--1464, 1953.

\bibitem[Conway and Sloane(2013)]{conway2013sphere}
John~Horton Conway and Neil James~Alexander Sloane.
\newblock \emph{Sphere packings, lattices and groups}, volume 290.
\newblock Springer Science \& Business Media, 2013.

\bibitem[Daskalakis et~al.(2021)Daskalakis, Skoulakis, and Zampetakis]{daskalakis2021complexity}
Constantinos Daskalakis, Stratis Skoulakis, and Manolis Zampetakis.
\newblock The complexity of constrained min-max optimization.
\newblock In \emph{Proceedings of the 53rd Annual ACM SIGACT Symposium on Theory of Computing}, pages 1466--1478, 2021.

\bibitem[Delsarte et~al.(1991)Delsarte, Goethals, and Seidel]{delsarte1991spherical}
Philippe Delsarte, Jean-Marie Goethals, and Johan~Jacob Seidel.
\newblock Spherical codes and designs.
\newblock In \emph{Geometry and Combinatorics}, pages 68--93. Elsevier, 1991.

\bibitem[Demircigil et~al.(2017)Demircigil, Heusel, L{\"o}we, Upgang, and Vermet]{demircigil2017model}
Mete Demircigil, Judith Heusel, Matthias L{\"o}we, Sven Upgang, and Franck Vermet.
\newblock On a model of associative memory with huge storage capacity.
\newblock \emph{Journal of Statistical Physics}, 168:\penalty0 288--299, 2017.

\bibitem[Devlin(2018)]{devlin2018bert}
Jacob Devlin.
\newblock Bert: Pre-training of deep bidirectional transformers for language understanding.
\newblock \emph{arXiv preprint arXiv:1810.04805}, 2018.

\bibitem[Dirksen et~al.(2022)Dirksen, Genzel, Jacques, and Stollenwerk]{JMLR:v23:21-1079}
Sjoerd Dirksen, Martin Genzel, Laurent Jacques, and Alexander Stollenwerk.
\newblock The separation capacity of random neural networks.
\newblock \emph{Journal of Machine Learning Research}, 23\penalty0 (309):\penalty0 1--47, 2022.

\bibitem[Fern{\'a}ndez et~al.(2021)Fern{\'a}ndez, Kim, Liu, and Pikhurko]{fernandez2021new}
Irene~Gil Fern{\'a}ndez, Jaehoon Kim, Hong Liu, and Oleg Pikhurko.
\newblock New lower bounds on kissing numbers and spherical codes in high dimensions.
\newblock \emph{arXiv preprint arXiv:2111.01255}, 2021.

\bibitem[F{\"u}rst et~al.(2022)F{\"u}rst, Rumetshofer, Lehner, Tran, Tang, Ramsauer, Kreil, Kopp, Klambauer, Bitto, et~al.]{furst2022cloob}
Andreas F{\"u}rst, Elisabeth Rumetshofer, Johannes Lehner, Viet~T Tran, Fei Tang, Hubert Ramsauer, David Kreil, Michael Kopp, G{\"u}nter Klambauer, Angela Bitto, et~al.
\newblock Cloob: Modern hopfield networks with infoloob outperform clip.
\newblock \emph{Advances in neural information processing systems (NeurIPS)}, 35:\penalty0 20450--20468, 2022.

\bibitem[Ghosal et~al.(2022)Ghosal, Mahankali, and Sun]{ghosal2022randomly}
Promit Ghosal, Srinath Mahankali, and Yihang Sun.
\newblock Randomly initialized one-layer neural networks make data linearly separable.
\newblock \emph{arXiv preprint arXiv:2205.11716}, 2022.

\bibitem[Graf et~al.(2021)Graf, Hofer, Niethammer, and Kwitt]{graf2021dissecting}
Florian Graf, Christoph Hofer, Marc Niethammer, and Roland Kwitt.
\newblock Dissecting supervised contrastive learning.
\newblock In \emph{International Conference on Machine Learning}, pages 3821--3830. PMLR, 2021.

\bibitem[He et~al.(2024)He, Karlinsky, Kim, McAuley, Krotov, and Feris]{he2024camelot}
Zexue He, Leonid Karlinsky, Donghyun Kim, Julian McAuley, Dmitry Krotov, and Rogerio Feris.
\newblock Camelot: Towards large language models with training-free consolidated associative memory.
\newblock \emph{arXiv preprint arXiv:2402.13449}, 2024.

\bibitem[Hofmann et~al.(2024)Hofmann, Schmid, Lehner, Klotz, and Hochreiter]{hofmann2024energy}
Claus Hofmann, Simon Schmid, Bernhard Lehner, Daniel Klotz, and Sepp Hochreiter.
\newblock Energy-based hopfield boosting for out-of-distribution detection.
\newblock \emph{arXiv preprint arXiv:2405.08766}, 2024.

\bibitem[Hoover et~al.(2023{\natexlab{a}})Hoover, Liang, Pham, Panda, Strobelt, Chau, Zaki, and Krotov]{hoover2023energy}
Benjamin Hoover, Yuchen Liang, Bao Pham, Rameswar Panda, Hendrik Strobelt, Duen~Horng Chau, Mohammed~J Zaki, and Dmitry Krotov.
\newblock Energy transformer.
\newblock \emph{arXiv preprint arXiv:2302.07253}, 2023{\natexlab{a}}.

\bibitem[Hoover et~al.(2023{\natexlab{b}})Hoover, Strobelt, Krotov, Hoffman, Kira, and Chau]{hoover2023memory}
Benjamin Hoover, Hendrik Strobelt, Dmitry Krotov, Judy Hoffman, Zsolt Kira, and Duen~Horng Chau.
\newblock Memory in plain sight: A survey of the uncanny resemblances between diffusion models and associative memories.
\newblock \emph{arXiv preprint arXiv:2309.16750}, 2023{\natexlab{b}}.

\bibitem[Hopfield(1982)]{hopfield1982neural}
John~J Hopfield.
\newblock Neural networks and physical systems with emergent collective computational abilities.
\newblock \emph{Proceedings of the national academy of sciences}, 79\penalty0 (8):\penalty0 2554--2558, 1982.

\bibitem[Hopfield(1984)]{hopfield1984neurons}
John~J Hopfield.
\newblock Neurons with graded response have collective computational properties like those of two-state neurons.
\newblock \emph{Proceedings of the national academy of sciences}, 81\penalty0 (10):\penalty0 3088--3092, 1984.

\bibitem[Hsieh et~al.(2021)Hsieh, Mertikopoulos, and Cevher]{hsieh2021limits}
Ya-Ping Hsieh, Panayotis Mertikopoulos, and Volkan Cevher.
\newblock The limits of min-max optimization algorithms: Convergence to spurious non-critical sets.
\newblock In \emph{International Conference on Machine Learning}, pages 4337--4348. PMLR, 2021.

\bibitem[Hsu(2012)]{hsu2012application}
Wei-Yen Hsu.
\newblock Application of competitive hopfield neural network to brain-computer interface systems.
\newblock \emph{International journal of neural systems}, 22\penalty0 (01):\penalty0 51--62, 2012.

\bibitem[Hu et~al.(2023)Hu, Yang, Wu, Xu, Chen, and Liu]{hu2023SparseHopfield}
Jerry Yao-Chieh Hu, Donglin Yang, Dennis Wu, Chenwei Xu, Bo-Yu Chen, and Han Liu.
\newblock On sparse modern hopfield model.
\newblock In \emph{Thirty-seventh Conference on Neural Information Processing Systems (NeurIPS)}, 2023.

\bibitem[Hu et~al.(2024{\natexlab{a}})Hu, Chang, Luo, Chen, Li, Wang, and Liu]{hu2024outlier}
Jerry Yao-Chieh Hu, Pei-Hsuan Chang, Haozheng Luo, Hong-Yu Chen, Weijian Li, Wei-Po Wang, and Han Liu.
\newblock Outlier-efficient hopfield layers for large transformer-based models.
\newblock In \emph{Forty-first International Conference on Machine Learning (ICML)}, 2024{\natexlab{a}}.

\bibitem[Hu et~al.(2024{\natexlab{b}})Hu, Chen, Wu, Ruan, and Liu]{hu2024nonparametric}
Jerry Yao-Chieh Hu, Bo-Yu Chen, Dennis Wu, Feng Ruan, and Han Liu.
\newblock Nonparametric modern hopfield models.
\newblock \emph{arXiv preprint arXiv:2404.03900}, 2024{\natexlab{b}}.

\bibitem[Hu et~al.(2024{\natexlab{c}})Hu, Lin, Song, and Liu]{hu2024computational}
Jerry Yao-Chieh Hu, Thomas Lin, Zhao Song, and Han Liu.
\newblock On computational limits of modern hopfield models: A fine-grained complexity analysis.
\newblock In \emph{Forty-first International Conference on Machine Learning (ICML)}, 2024{\natexlab{c}}.

\bibitem[Hu et~al.(2024{\natexlab{d}})Hu, Wu, Li, Pi, , Song, and Liu]{hwsl24}
Jerry Yao-Chieh Hu, Weimin Wu, Zhuoru Li, Sophia Pi, , Zhao Song, and Han Liu.
\newblock On statistical rates and provably efficient criteria of latent diffusion transformers (dits).
\newblock In \emph{Thirty-eighth Conference on Neural Information Processing Systems (NeurIPS)}, 2024{\natexlab{d}}.

\bibitem[Iatropoulos et~al.(2022)Iatropoulos, Brea, and Gerstner]{iatropoulos2022kernel}
Georgios Iatropoulos, Johanni Brea, and Wulfram Gerstner.
\newblock Kernel memory networks: A unifying framework for memory modeling.
\newblock \emph{Advances in Neural Information Processing Systems}, 35:\penalty0 35326--35338, 2022.

\bibitem[Iusem(2003)]{iusem2003convergence}
Alfredo~N Iusem.
\newblock On the convergence properties of the projected gradient method for convex optimization.
\newblock \emph{Computational \& Applied Mathematics}, 22:\penalty0 37--52, 2003.

\bibitem[Jenssen et~al.(2018)Jenssen, Joos, and Perkins]{jenssen2018kissing}
Matthew Jenssen, Felix Joos, and Will Perkins.
\newblock On kissing numbers and spherical codes in high dimensions.
\newblock \emph{Advances in Mathematics}, 335:\penalty0 307--321, 2018.

\bibitem[Ji et~al.(2021)Ji, Zhou, Liu, and Davuluri]{ji2021dnabert}
Yanrong Ji, Zhihan Zhou, Han Liu, and Ramana~V Davuluri.
\newblock Dnabert: pre-trained bidirectional encoder representations from transformers model for dna-language in genome.
\newblock \emph{Bioinformatics}, 37\penalty0 (15):\penalty0 2112--2120, 2021.

\bibitem[Jiang et~al.(2023)Jiang, Zhou, Wang, Qu, Mixon, You, and Zhu]{jiang2023generalized}
Jiachen Jiang, Jinxin Zhou, Peng Wang, Qing Qu, Dustin Mixon, Chong You, and Zhihui Zhu.
\newblock Generalized neural collapse for a large number of classes.
\newblock \emph{arXiv preprint arXiv:2310.05351}, 2023.

\bibitem[Kabatiansky and Levenshtein(1978)]{kabatiansky1978bounds}
Grigorii~Anatol'evich Kabatiansky and Vladimir~Iosifovich Levenshtein.
\newblock On bounds for packings on a sphere and in space.
\newblock \emph{Problemy peredachi informatsii}, 14\penalty0 (1):\penalty0 3--25, 1978.

\bibitem[Kanerva(1988)]{kanerva1988sparse}
Pentti Kanerva.
\newblock \emph{Sparse distributed memory}.
\newblock MIT press, 1988.

\bibitem[Kojima et~al.(2022)Kojima, Gu, Reid, Matsuo, and Iwasawa]{kojima2022large}
Takeshi Kojima, Shixiang~Shane Gu, Machel Reid, Yutaka Matsuo, and Yusuke Iwasawa.
\newblock Large language models are zero-shot reasoners.
\newblock \emph{Advances in neural information processing systems}, 35:\penalty0 22199--22213, 2022.

\bibitem[Kozachkov et~al.(2023)Kozachkov, Kastanenka, and Krotov]{kozachkov2023building}
Leo Kozachkov, Ksenia~V Kastanenka, and Dmitry Krotov.
\newblock Building transformers from neurons and astrocytes.
\newblock \emph{Proceedings of the National Academy of Sciences}, 120\penalty0 (34):\penalty0 e2219150120, 2023.

\bibitem[Krotov(2023)]{krotov2023new}
Dmitry Krotov.
\newblock A new frontier for hopfield networks.
\newblock \emph{Nature Reviews Physics}, 5\penalty0 (7):\penalty0 366--367, 2023.

\bibitem[Krotov and Hopfield(2016)]{krotov2016dense}
Dmitry Krotov and John~J Hopfield.
\newblock Dense associative memory for pattern recognition.
\newblock \emph{Advances in neural information processing systems}, 29, 2016.

\bibitem[Krotov and Hopfield(2021)]{krotov2020large}
Dmitry Krotov and John~J. Hopfield.
\newblock Large associative memory problem in neurobiology and machine learning.
\newblock In \emph{International Conference on Learning Representations (ICLR)}, 2021.

\bibitem[Lebedev and Nicolelis(2006)]{lebedev2006brain}
Mikhail~A Lebedev and Miguel~AL Nicolelis.
\newblock Brain--machine interfaces: past, present and future.
\newblock \emph{TRENDS in Neurosciences}, 29\penalty0 (9):\penalty0 536--546, 2006.

\bibitem[L{\'e}cuyer et~al.(2008)L{\'e}cuyer, Lotte, Reilly, Leeb, Hirose, and Slater]{lecuyer2008brain}
Anatole L{\'e}cuyer, Fabien Lotte, Richard~B Reilly, Robert Leeb, Michitaka Hirose, and Mel Slater.
\newblock Brain-computer interfaces, virtual reality, and videogames.
\newblock \emph{Computer}, 41\penalty0 (10):\penalty0 66--72, 2008.

\bibitem[Lelewer and Hirschberg(1987)]{lelewer1987data}
Debra~A Lelewer and Daniel~S Hirschberg.
\newblock Data compression.
\newblock \emph{ACM Computing Surveys (CSUR)}, 19\penalty0 (3):\penalty0 261--296, 1987.

\bibitem[Liu et~al.(2015)Liu, Zhang, Subei, Richardson, Lucas, and Van~der Spiegel]{liu2015pennbmbi}
Xilin Liu, Milin Zhang, Basheer Subei, Andrew~G Richardson, Timothy~H Lucas, and Jan Van~der Spiegel.
\newblock The pennbmbi: Design of a general purpose wireless brain-machine-brain interface system.
\newblock \emph{IEEE transactions on biomedical circuits and systems}, 9\penalty0 (2):\penalty0 248--258, 2015.

\bibitem[Lucibello and M{\'e}zard(2024)]{lucibello2024exponential}
Carlo Lucibello and Marc M{\'e}zard.
\newblock Exponential capacity of dense associative memories.
\newblock \emph{Physical Review Letters}, 132\penalty0 (7):\penalty0 077301, 2024.

\bibitem[Martins et~al.(2023)Martins, Niculae, and McNamee]{martins2023sparse}
Andre Martins, Vlad Niculae, and Daniel~C McNamee.
\newblock Sparse modern hopfield networks.
\newblock In \emph{Associative Memory {\&} Hopfield Networks in 2023}, 2023.

\bibitem[Moore(1974)]{moore1974vector}
Michael~H Moore.
\newblock Vector packing in finite dimensional vector spaces.
\newblock \emph{Linear Algebra and its Applications}, 8\penalty0 (3):\penalty0 213--224, 1974.

\bibitem[Morrison et~al.(2017)Morrison, Maia, Kutz, et~al.]{morrison2017preventing}
Megan Morrison, Pedro~D Maia, J~Nathan Kutz, et~al.
\newblock Preventing neurodegenerative memory loss in hopfield neuronal networks using cerebral organoids or external microelectronics.
\newblock \emph{Computational and Mathematical Methods in Medicine}, 2017, 2017.

\bibitem[Neelakantan et~al.(2022)Neelakantan, Xu, Puri, Radford, Han, Tworek, Yuan, Tezak, Kim, Hallacy, et~al.]{neelakantan2022text}
Arvind Neelakantan, Tao Xu, Raul Puri, Alec Radford, Jesse~Michael Han, Jerry Tworek, Qiming Yuan, Nikolas Tezak, Jong~Wook Kim, Chris Hallacy, et~al.
\newblock Text and code embeddings by contrastive pre-training.
\newblock \emph{arXiv preprint arXiv:2201.10005}, 2022.

\bibitem[OpenAI(2024)]{SoRA_2024}
OpenAI.
\newblock Sora: A video generative model based on transformer diffusion.
\newblock \emph{OpenAI Research}, 2024.
\newblock Accessed: 08/16/2024.

\bibitem[Ota et~al.(2023)Ota, Sato, Kawakami, Tanaka, and Inoue]{ota2023learning}
Toshihiro Ota, Ikuro Sato, Rei Kawakami, Masayuki Tanaka, and Nakamasa Inoue.
\newblock Learning with partial forgetting in modern hopfield networks.
\newblock In \emph{International Conference on Artificial Intelligence and Statistics (AISTATS)}, pages 6661--6673. PMLR, 2023.

\bibitem[Papyan et~al.(2020)Papyan, Han, and Donoho]{papyan2020prevalence}
Vardan Papyan, XY~Han, and David~L Donoho.
\newblock Prevalence of neural collapse during the terminal phase of deep learning training.
\newblock \emph{Proceedings of the National Academy of Sciences}, 117\penalty0 (40):\penalty0 24652--24663, 2020.

\bibitem[Peebles and Xie(2023)]{peebles2023scalable}
William Peebles and Saining Xie.
\newblock Scalable diffusion models with transformers.
\newblock In \emph{Proceedings of the IEEE/CVF International Conference on Computer Vision (ICCV)}, pages 4195--4205, 2023.

\bibitem[Peters et~al.(2019)Peters, Niculae, and Martins]{peters2019sparse}
Ben Peters, Vlad Niculae, and Andr{\'e}~FT Martins.
\newblock Sparse sequence-to-sequence models.
\newblock \emph{arXiv preprint arXiv:1905.05702}, 2019.

\bibitem[Peterson and Weldon(1972)]{peterson1972error}
William~Wesley Peterson and Edward~J Weldon.
\newblock \emph{Error-correcting codes}.
\newblock MIT press, 1972.

\bibitem[Polyak et~al.(2024)Polyak, Zohar, Brown, Tjandra, Sinha, Lee, Vyas, Shi, Ma, Chuang, et~al.]{polyak2024movie}
Adam Polyak, Amit Zohar, Andrew Brown, Andros Tjandra, Animesh Sinha, Ann Lee, Apoorv Vyas, Bowen Shi, Chih-Yao Ma, Ching-Yao Chuang, et~al.
\newblock Movie gen: A cast of media foundation models.
\newblock \emph{arXiv preprint arXiv:2410.13720}, 2024.

\bibitem[Polyak and Juditsky(1992)]{polyak1992acceleration}
Boris~T Polyak and Anatoli~B Juditsky.
\newblock Acceleration of stochastic approximation by averaging.
\newblock \emph{SIAM journal on control and optimization}, 30\penalty0 (4):\penalty0 838--855, 1992.

\bibitem[Radford et~al.(2019)Radford, Wu, Child, Luan, Amodei, Sutskever, et~al.]{radford2019language}
Alec Radford, Jeffrey Wu, Rewon Child, David Luan, Dario Amodei, Ilya Sutskever, et~al.
\newblock Language models are unsupervised multitask learners.
\newblock \emph{OpenAI blog}, 1\penalty0 (8):\penalty0 9, 2019.

\bibitem[Raman and Yang(2019)]{raman2019optimization}
Parameswaran Raman and Jiasen Yang.
\newblock Optimization on the surface of the (hyper)-sphere.
\newblock \emph{arXiv preprint arXiv:1909.06463}, 2019.

\bibitem[Ramos-Murguialday et~al.(2013)Ramos-Murguialday, Broetz, Rea, L{\"a}er, Yilmaz, Brasil, Liberati, Curado, Garcia-Cossio, Vyziotis, et~al.]{ramos2013brain}
Ander Ramos-Murguialday, Doris Broetz, Massimiliano Rea, Leonhard L{\"a}er, {\"O}zge Yilmaz, Fabricio~L Brasil, Giulia Liberati, Marco~R Curado, Eliana Garcia-Cossio, Alexandros Vyziotis, et~al.
\newblock Brain--machine interface in chronic stroke rehabilitation: a controlled study.
\newblock \emph{Annals of neurology}, 74\penalty0 (1):\penalty0 100--108, 2013.

\bibitem[Ramsauer et~al.(2020)Ramsauer, Sch{\"a}fl, Lehner, Seidl, Widrich, Adler, Gruber, Holzleitner, Pavlovi{\'c}, Sandve, et~al.]{ramsauer2020hopfield}
Hubert Ramsauer, Bernhard Sch{\"a}fl, Johannes Lehner, Philipp Seidl, Michael Widrich, Thomas Adler, Lukas Gruber, Markus Holzleitner, Milena Pavlovi{\'c}, Geir~Kjetil Sandve, et~al.
\newblock Hopfield networks is all you need.
\newblock \emph{arXiv preprint arXiv:2008.02217}, 2020.

\bibitem[Rockafellar and Wets(2009)]{rockafellar2009variational}
R~Tyrrell Rockafellar and Roger J-B Wets.
\newblock \emph{Variational analysis}, volume 317.
\newblock Springer Science \& Business Media, 2009.

\bibitem[Saha et~al.(2023)Saha, Krotov, Zaki, and Ram]{saha2023end}
Bishwajit Saha, Dmitry Krotov, Mohammed~J Zaki, and Parikshit Ram.
\newblock End-to-end differentiable clustering with associative memories.
\newblock In \emph{International Conference on Machine Learning}, pages 29649--29670. PMLR, 2023.

\bibitem[Santos et~al.(2024)Santos, Niculae, McNamee, and Martins]{santos2024sparse}
Saul Santos, Vlad Niculae, Daniel McNamee, and Andre~FT Martins.
\newblock Sparse and structured hopfield networks.
\newblock \emph{arXiv preprint arXiv:2402.13725}, 2024.

\bibitem[Seidl et~al.(2022)Seidl, Renz, Dyubankova, Neves, Verhoeven, Wegner, Segler, Hochreiter, and Klambauer]{seidl2022improving}
Philipp Seidl, Philipp Renz, Natalia Dyubankova, Paulo Neves, Jonas Verhoeven, Jorg~K Wegner, Marwin Segler, Sepp Hochreiter, and Gunter Klambauer.
\newblock Improving few-and zero-shot reaction template prediction using modern hopfield networks.
\newblock \emph{Journal of chemical information and modeling}, 62\penalty0 (9):\penalty0 2111--2120, 2022.

\bibitem[Shanechi(2019)]{shanechi2019brain}
Maryam~M Shanechi.
\newblock Brain--machine interfaces from motor to mood.
\newblock \emph{Nature neuroscience}, 22\penalty0 (10):\penalty0 1554--1564, 2019.

\bibitem[Shannon(1959)]{shannon1959probability}
Claude~E Shannon.
\newblock Probability of error for optimal codes in a gaussian channel.
\newblock \emph{Bell System Technical Journal}, 38\penalty0 (3):\penalty0 611--656, 1959.

\bibitem[Shen et~al.(2020)Shen, Chen, Heaton, Chen, Liu, Yin, and Wang]{shen2020learning}
Jiayi Shen, Xiaohan Chen, Howard Heaton, Tianlong Chen, Jialin Liu, Wotao Yin, and Zhangyang Wang.
\newblock Learning a minimax optimizer: A pilot study.
\newblock In \emph{International Conference on Learning Representations}, 2020.

\bibitem[Sriperumbudur and Lanckriet(2009)]{sriperumbudur2009convergence}
Bharath~K Sriperumbudur and Gert~RG Lanckriet.
\newblock On the convergence of the concave-convex procedure.
\newblock In \emph{Advances in neural information processing systems}, volume~9, pages 1759--1767, 2009.

\bibitem[Strohmer and Heath~Jr(2003)]{strohmer2003grassmannian}
Thomas Strohmer and Robert~W Heath~Jr.
\newblock Grassmannian frames with applications to coding and communication.
\newblock \emph{Applied and computational harmonic analysis}, 14\penalty0 (3):\penalty0 257--275, 2003.

\bibitem[Touvron et~al.(2023)Touvron, Lavril, Izacard, Martinet, Lachaux, Lacroix, Rozi{\`e}re, Goyal, Hambro, Azhar, et~al.]{touvron2023llama}
Hugo Touvron, Thibaut Lavril, Gautier Izacard, Xavier Martinet, Marie-Anne Lachaux, Timoth{\'e}e Lacroix, Baptiste Rozi{\`e}re, Naman Goyal, Eric Hambro, Faisal Azhar, et~al.
\newblock Llama: Open and efficient foundation language models.
\newblock \emph{arXiv preprint arXiv:2302.13971}, 2023.

\bibitem[Tripuraneni et~al.(2018)Tripuraneni, Flammarion, Bach, and Jordan]{tripuraneni2018averaging}
Nilesh Tripuraneni, Nicolas Flammarion, Francis Bach, and Michael~I Jordan.
\newblock Averaging stochastic gradient descent on riemannian manifolds.
\newblock In \emph{Conference On Learning Theory}, pages 650--687. PMLR, 2018.

\bibitem[Vaswani et~al.(2017)Vaswani, Shazeer, Parmar, Uszkoreit, Jones, Gomez, Kaiser, and Polosukhin]{vaswani2017attention}
Ashish Vaswani, Noam Shazeer, Niki Parmar, Jakob Uszkoreit, Llion Jones, Aidan~N Gomez, {\L}ukasz Kaiser, and Illia Polosukhin.
\newblock Attention is all you need.
\newblock \emph{Advances in neural information processing systems}, 30, 2017.

\bibitem[Wang(2009)]{wang2009finding}
Jeffrey Wang.
\newblock Finding and investigating exact spherical codes.
\newblock \emph{Experimental Mathematics}, 18\penalty0 (2):\penalty0 249--256, 2009.

\bibitem[Wang et~al.(2023)Wang, Yang, Huang, Yang, Majumder, and Wei]{wang2023improving}
Liang Wang, Nan Yang, Xiaolong Huang, Linjun Yang, Rangan Majumder, and Furu Wei.
\newblock Improving text embeddings with large language models.
\newblock \emph{arXiv preprint arXiv:2401.00368}, 2023.

\bibitem[Widrich et~al.(2020)Widrich, Sch{\"a}fl, Pavlovi{\'c}, Ramsauer, Gruber, Holzleitner, Brandstetter, Sandve, Greiff, Hochreiter, et~al.]{widrich2020modern}
Michael Widrich, Bernhard Sch{\"a}fl, Milena Pavlovi{\'c}, Hubert Ramsauer, Lukas Gruber, Markus Holzleitner, Johannes Brandstetter, Geir~Kjetil Sandve, Victor Greiff, Sepp Hochreiter, et~al.
\newblock Modern hopfield networks and attention for immune repertoire classification.
\newblock \emph{Advances in Neural Information Processing Systems}, 33:\penalty0 18832--18845, 2020.

\bibitem[Willshaw et~al.(1969)Willshaw, Buneman, and Longuet-Higgins]{willshaw1969non}
David~J Willshaw, O~Peter Buneman, and Hugh~Christopher Longuet-Higgins.
\newblock Non-holographic associative memory.
\newblock \emph{Nature}, 222\penalty0 (5197):\penalty0 960--962, 1969.

\bibitem[Wu et~al.(2024{\natexlab{a}})Wu, Hu, Hsiao, and Liu]{wu2024uniform}
Dennis Wu, Jerry Yao-Chieh Hu, Teng-Yun Hsiao, and Han Liu.
\newblock Uniform memory retrieval with larger capacity for modern hopfield models.
\newblock In \emph{Forty-first International Conference on Machine Learning (ICML)}, 2024{\natexlab{a}}.

\bibitem[Wu et~al.(2024{\natexlab{b}})Wu, Hu, Li, Chen, and Liu]{wu2023stanhop}
Dennis Wu, Jerry Yao-Chieh Hu, Weijian Li, Bo-Yu Chen, and Han Liu.
\newblock Stanhop: Sparse tandem hopfield model for memory-enhanced time series prediction.
\newblock In \emph{The Twelfth International Conference on Learning Representations (ICLR)}, 2024{\natexlab{b}}.

\bibitem[Wyner(1965)]{wyner1965capabilities}
Aaron~D Wyner.
\newblock Capabilities of bounded discrepancy decoding.
\newblock \emph{Bell System Technical Journal}, 44\penalty0 (6):\penalty0 1061--1122, 1965.

\bibitem[Xu et~al.(2024)Xu, Huang, Hu, Li, Gilani, Goan, and Liu]{xu2024bishop}
Chenwei Xu, Yu-Chao Huang, Jerry Yao-Chieh Hu, Weijian Li, Ammar Gilani, Hsi-Sheng Goan, and Han Liu.
\newblock Bishop: Bi-directional cellular learning for tabular data with generalized sparse modern hopfield model.
\newblock \emph{arXiv preprint arXiv:2404.03830}, 2024.

\bibitem[Yuille and Rangarajan(2001)]{yuille2001concave}
Alan~L Yuille and Anand Rangarajan.
\newblock The concave-convex procedure (cccp).
\newblock \emph{Advances in neural information processing systems}, 14, 2001.

\bibitem[Zhou et~al.(2023)Zhou, Ji, Li, Dutta, Davuluri, and Liu]{zhou2023dnabert}
Zhihan Zhou, Yanrong Ji, Weijian Li, Pratik Dutta, Ramana Davuluri, and Han Liu.
\newblock Dnabert-2: Efficient foundation model and benchmark for multi-species genome.
\newblock \emph{arXiv preprint arXiv:2306.15006}, 2023.

\bibitem[Zhou et~al.(2024)Zhou, Wu, Ho, Wang, Shi, Davuluri, Wang, and Liu]{zhou2024dnabertS}
Zhihan Zhou, Winmin Wu, Harrison Ho, Jiayi Wang, Lizhen Shi, Ramana~V Davuluri, Zhong Wang, and Han Liu.
\newblock Dnabert-s: Learning species-aware dna embedding with genome foundation models.
\newblock \emph{arXiv preprint arXiv:2402.08777}, 2024.

\bibitem[Zhu et~al.(2021)Zhu, Ding, Zhou, Li, You, Sulam, and Qu]{zhu2021geometric}
Zhihui Zhu, Tianyu Ding, Jinxin Zhou, Xiao Li, Chong You, Jeremias Sulam, and Qing Qu.
\newblock A geometric analysis of neural collapse with unconstrained features.
\newblock \emph{Advances in Neural Information Processing Systems}, 34:\penalty0 29820--29834, 2021.

\end{thebibliography}

\newpage

\setlength{\abovedisplayskip}{10pt}
\setlength{\abovedisplayshortskip}{10pt}
\setlength{\belowdisplayskip}{10pt}
\setlength{\belowdisplayshortskip}{10pt}

\setlist[itemize]{leftmargin=1em}
\setlist[enumerate]{leftmargin=1.4em}

\appendix
\label{sec:append}
\part*{Appendix}
{
\setlength{\parskip}{-0em}
\startcontents[sections]
\printcontents[sections]{ }{1}{}
}

{
\setlength{\parskip}{-0em}
\startcontents[sections]
\printcontents[sections]{ }{1}{}
}

\clearpage
\section{Related Works}
\label{sec:related_work}
\paragraph{Hopfield and Dense Associative Memory Models.}

Associative memory models \cite{kanerva1988sparse, willshaw1969non} are extensively studied in neuroscience and machine learning due to their biologically plausible designs. 
These models aim to store a set of memories and accurately retrieve each one given an input query. 
Hopfield models \cite{hopfield1982neural} are a class of energy-based associative memory models, beginning with classical versions \cite{hopfield1984neurons, hopfield1982neural} that handle binary patterns and have a (sub-)linear memory capacity of $\calO(d)$ for a pattern dimension $d$. 
Dense associative memory models \cite{krotov2020large,krotov2016dense, ramsauer2020hopfield, demircigil2017model} are later proposed with superlinear memory capacity enabled by sharper energy functions (e.g., polynomial and exponential). 
Notably, the latest advancement in these dense models is their exponential-in-$d$ capacity \cite{lucibello2024exponential,santos2024sparse,wu2023stanhop, hu2024outlier, hu2024computational, hu2024nonparametric,hu2023SparseHopfield, ramsauer2020hopfield}.
However, the current literature reports are mostly\footnote{After completion of this work, the authors attended ICML 2024 and learned of an upper bound result reported by \cite{santos2024sparse}, which also utilizes techniques from spherical codes. 
The authors regret missing the dinner with \citet{santos2024sparse} at Vienna due to a tight schedule.
The difference between this work and \cite{santos2024sparse} lies not only in scope but also in that our upper bound matches the lower bound in the low-temperature region (\cref{prop:capacity-bound}).} lower bound results for this exponential memory capacity. 
This work matches these lower bounds with an upper bound, making the capacity both tight and provably optimal.

\paragraph{Transformer-Compatible Dense Associative Memories: Modern Hopfield Models.} 
Recently, a special class of dense associative memory models, modern Hopfield models (MHMs), has gained increasing interest in deep learning due to their connection to the attention mechanism in transformers \cite{wu2024uniform,wu2023stanhop,hu2024outlier,hu2024computational,hu2024nonparametric,hu2023SparseHopfield,ramsauer2020hopfield}. 
Therefore, we also refer to them as \textit{transformer-compatible} dense associative memories.
Notably, the defining characteristic of MHMs is that their single-step update is equivalent to the attention mechanism \cite{vaswani2017attention}.
This striking feature makes them interesting given the prevalence and dominance of transformer architectures in the era of large foundation models \cite{polyak2024movie,SoRA_2024,hwsl24,peebles2023scalable,bubeck2023sparks,bai2023transformers,achiam2023gpt,zhou2024dnabertS,zhou2023dnabert,ji2021dnabert,touvron2023llama,kojima2022large,bommasani2021opportunities,radford2019language,devlin2018bert}.

As a result, such a connection facilitates the integration of associative memory models into modern deep learning and large foundation models 
\cite{burns2024semantically, xu2024bishop, hu2024outlier, hu2024nonparametric, hofmann2024energy, wu2023stanhop, auer2023conformal, burns2023simplicial, furst2022cloob, krotov2016dense}.
Moreover, recent studies introduce several modern Hopfield model variants.
\citet{hu2023SparseHopfield} propose the sparse modern Hopfield model, a sparse counterpart to MHM with larger capacity and lower retrieval error.
\citet{wu2023stanhop,santos2024sparse,martins2023sparse} introduce generalized sparse Hopfield models that unify all MHMs with different degrees of sparsity.
\citet{hu2024nonparametric} present a nonparametric construction for deep learning compatible Hopfield layers and several efficient modern Hopfield variants. 
\citet{hu2024outlier} introduce OutEffHop, an outlier-removing, deep learning compatible Hopfield layer for robust large pretrained model quantization.
\citet{wu2024uniform} propose $\mathtt{U}\text{-}\mathtt{Hop}$ facilitating memory retrieval in a learnable feature space (see introduce for a review).
Empirically, $\mathtt{U}\text{-}\mathtt{Hop}$   improves the memory confusion problem \cite{krotov2020large} by a significant margin.

\paragraph{Applications of Modern Hopfield Models in Modern Machine Learning.} 
Recently, modern Hopfield models also achieve empirical success across various deep learning tasks \cite{krotov2023new}. 
Starting with \cite{krotov2016dense}, the polynomial Hopfield model is proposed for image classification tasks (e.g., MNIST). 
Later, \citet{ramsauer2020hopfield} introduce modern Hopfield layers compatible with deep learning architectures. 
Since then, various modern Hopfield layers \cite{hu2024outlier, wu2023stanhop, hu2023SparseHopfield} have been applied in many deep learning tasks, such as language modeling \cite{he2024camelot}, multiple instance learning \cite{ramsauer2020hopfield}, immune repertoire classification \cite{widrich2020modern}, multivariate time series prediction \cite{wu2023stanhop}, image generation \cite{hoover2023memory}, tabular learning \cite{xu2024bishop}, unsupervised clustering \cite{saha2023end}, and image captioning \cite{furst2022cloob}.

Additionally, modern Hopfield layers  introduce new operations in large foundation models to improve performance.
For example, \citet{wu2023stanhop} leverage the retrieval dynamics of modern Hopfield models to enable an external memory plugin in time series prediction, and \citet{xu2024bishop} apply a similar approach to tabular data. 
In image captioning, \citet{furst2022cloob} address issues with covariance structure using modern Hopfield models. 
\citet{hu2024outlier} propose an outlier-free Hopfield layer as a quantization-strong and resource-efficient transformer backbone for large language models and large foundation models. 
\citet{ota2023learning} embed a partial forgetting functionality in modern Hopfield models to enhance model performance.

\paragraph{Kernelized Hopfield Models.}
\citet{wu2024uniform} propose kernelized Hopfield models (KHMs)\footnote{This is different from \cite{iatropoulos2022kernel} while they share similar names.}, which have the capability to store memories in a learnable feature space. 
KHMs offer the flexibility to relocate memories while maintaining several defining properties of modern Hopfield models \cite[Theorems 1 to 4]{ramsauer2020hopfield}. 
By maximizing the average separation between memories, KHMs empirically achieve lower retrieval errors \cite[Section 4]{wu2024uniform}. 
However, the theoretical understanding of KHMs is larking due to their new flexibility in rearranging memories. 
This work aims to fill this gap by 
analyzing the capacity limits and theoretical justifications for KHMs.

\paragraph{Spherical Code.}
Spherical codes are mathematical constructions describing the arrangement of $M$ points on the surface of a $d$-dimensional hyper-sphere \cite{delsarte1991spherical}.
The main problem around spherical codes is to arrange points in a way such that the minimum distance between any two points are maximized.
This arrangement is called the optimal spherical code.
It is crucial for minimizing errors and maximizing efficiency in signal transmission and data storage.
In general, the value of max minimal separation and arrangement of points is unsolved except for certain pairs of $(d, M)$ \cite{wang2009finding}.
Various of fields such as communications \cite{strohmer2003grassmannian} and data compression \cite{lelewer1987data}.
Moreover, spherical codes are related to the area of error-correcting codes \cite{peterson1972error}, which are used to detect and correct errors in data transmission and storage.

\section{More Discussions}

\subsection{MHM Capacity}\label{appendix:MHM-capacity-explained}
We review the modern Hopfield capacity lower bound in \cite{ramsauer2020hopfield}.
\begin{lemma}[Memory Capacity of MHM]
    Let $1-p$ be the probability of successfully storing
    and retrieving a pattern. 
    Assuming patterns are normalized, the amount of patterns randomly sampled from a $d$-dimensional unit-sphere that the MHM with update rule in \eqref{MHM-update}, can store and retrieve is lower-bounded by
    \begin{align*}
        M \geq \sqrt{p} C^{(d-1)/4},
    \end{align*}
    where $C$ is the solution to $C = \nicefrac{b}{W_0 \(\exp{a + \ln{b}}\)}$, with $W_0(\cdot)$ being the principal branch of Lambert $W$ function,
    $a \coloneqq (\nicefrac{4}{d-1}) \( \ln \nicefrac{2\sqrt{p}-2}{R} + 1\) $ and $b \coloneqq \nicefrac{4\beta}{5(d-1)}$.
\end{lemma}
Observe $W_0$ is an increasing function in $\exp{a + \ln b}$, indicating that $C$ is increasing in $R$ ( $C$ is increasing in $a$; $a$ is decreasing with fixed $R$).
Finally, we observe that under \cref{assumption1}, we have $R = \frac{1}{2} \sqrt{2\Delta_{\min}}$, implying that the memory capacity of MHM is constrained by large $\Delta_{\min}$.

\subsection{Learning on Stiefel Manifolds}\label{pgd}

We review ways to do optimization  on the surface of an unit-hypersphere (Stiefel manifold).

\paragraph{Projected Gradient Descent.}

Given any loss function $\calL(\cdot)$, an input matrix $\bX$,  and learning rate $\gamma$, a single gradient descent step at $t$-th time step is:
\begin{equation}\label{pgd1}
    \bW_{t+\nicefrac{1}{2}} = \bW_t - \gamma_t \nabla_\bX \calL(\bX).
\end{equation}
Projected gradient descent \cite{raman2019optimization} then projects $\bW_{t+\frac{1}{2}}$ onto the feasible set,  in this case, the surface of a unit hypersphere 
\begin{equation}\label{pgd2}
    \bW_{t+1} = \frac{\bW_{t+\nicefrac{1}{2}}}{ \norm{\bW_{t + \nicefrac{1}{2}}}}.
\end{equation}
Combining \eqref{pgd1} and \eqref{pgd2}, we obtain the projected gradient descent step as
\begin{align*}
    \bW_{t+1} = \mathtt{PGD}( \bW_t, \gamma, \bX).
\end{align*}

\paragraph{Riemannian Optimization.}

\cite{tripuraneni2018averaging} 
construct and analyze an approach on optimization on Riemannian manifolds from a geometric perspective.
Their adapt the \textit{Polyak-Ruppert} \cite{polyak1992acceleration} iterate averaging technique to the Riemannian setting. 
In general, with carefully selected step-size, their method achieves $\calO(\nicefrac{1}{N})$ convergence rate which is the same in the euclidean setting.
Overall, there are various of methods for optimization on Riemannian manifolds with comparable convergence rate to Euclidean space.
Thus, giving the advancement of Riemannian optimization, our assumption in \cref{def:function-class} is reasonably mild.

\paragraph{L2 Regularization.}
A simple alternative to satisfy the norm constraint is through L2 regularization.
From the aspect of neural collapse \cite{papyan2020prevalence}, learning under \eqref{eqn:sep-loss} is similar to learning under cross-entropy loss or supervised contrastive learning without positive samples \cite{graf2021dissecting}.
Further, from the analysis on unconstrainted feature models \cite{zhu2021geometric}, we can see that with carefully chosen coefficient on the regularization term on $\bW$, the optimal solution of \eqref{eqn:sep-loss} or cross-entropy loss ended up outputting normalized features.

Based on the previous research, we are able to see that the constraint in \cref{def:function-class} is not difficult to satisfy.

\subsection{KHMs as Brain-Machine Interface}

The ultimate goal in the field of Brain-Machine Interface (BMI) \cite{lebedev2006brain} is to design communication between human brains and external devices. It has potential in various real-world applications such as medical treatment \cite{ramos2013brain, shanechi2019brain}, virtual reality \cite{lecuyer2008brain}, etc. 
While Hopfield models serve as computational models for simulating human brains and their memory recall system \cite{liu2015pennbmbi, hsu2012application}, KHMs correspond to external devices (storage space) to assist/enhance the process of memory recall.

In this paper, we study the scenario where we try to optimize the usage of external neurons for memory storage by increasing the separation between memories in such external space ($\Phi$-space). 
We show that to prevent memory confusion, minimizing separation loss is an effective way to utilize the external space efficiently.
The closest work we can find is \cite{morrison2017preventing} and \cite{kozachkov2023building}.
In particular, they use Hopfield (and modern  Hopfield or dense associative memory) models as the computational model for human brain.
Specifically, \citet{morrison2017preventing} study the case of memory loss caused by disease or injuries.
Their proposed framework is able to consider the level of memory damage and then estimate the required dimension to fully or partially recover memories.
\citet{kozachkov2023building} explore the potential of building biological computers through links between transformers and modern Hopfield models. They show that neuron–astrocyte networks can perform the core computations of a transformer. In this context, our work provide an improved  connection between brain and transformer models, offering a more learnable and powerful approach to brain-machine interfaces (BMI).

\clearpage
\section{Proofs of Main Text}

\subsection{Proof of \texorpdfstring{\cref{lem:KHM-lower-bound}}{}}
\label{proof:capacity-lower-bound}

We first introduce a helper lemma:
\begin{lemma}[\cite{ramsauer2020hopfield, hu2023SparseHopfield}]\label{lem:technical1}
    Given real numbers, $a,b \in \R$. If the equation
    \begin{align*}
        ac + c \ln{c} - b = 0,
    \end{align*}
    holds, then the solution is
    \begin{align*}
        c = \frac{b}{ W_0 ( \exp{a + \ln{b}} ) }.
    \end{align*}
\end{lemma}

By \citep[Corollary~3.1.1]{hu2023SparseHopfield}, we state the well-separation condition of dense modern Hopfield model \cite{ramsauer2020hopfield}.
\begin{lemma}[Well Separation Condition of Dense Modern Hopfield Model \cite{ramsauer2020hopfield}]
\label{lemma:well_sep_dense}
    Following \cref{def:stored_and_retrieved}, suppose the memory patterns $\{ \bxi_\mu\}_{\mu\in[M]}$ are located within the sphere $S_\mu \coloneqq \{ \bx | \norm{\bx - \bxi_\mu} \leq R \}$.
    Then, assuming normalized memory patterns, the retrieval dynamics $\calT_{\text{MHM}}$ maps the sphere $S_\mu$ onto itself under the following conditions:
    \begin{enumerate}
        \item The initial query $\bx$ is located within the sphere $S_\mu$, i.e. $\bx \in S_\mu$.
        \item The \textit{well-separation} condition is satisfied, which is given by:
    \end{enumerate}    
    \begin{align*}
            \Delta_\mu \geq \frac{1}{\beta} \ln{\frac{ 2(M-1) }{R}} + 2R.
    \end{align*}
\end{lemma}
This specifies the necessary condition for a pattern $\bxi_\mu$ to be stored in $E$ and be able to retrieved by $\calT$.

\begin{proof}[Proof of \cref{lem:KHM-lower-bound}]

Let $ \Delta_{\min}^{\Phi} = \Min_{\mu\in[M]} \Delta_{\mu}^{\Phi}$, and $\theta_{\mu\nu, \Phi}$ be the angle between two patterns $\Phi(\bxi_\nu)$ and $\Phi(\bxi_\mu)$.
Note that $\theta_{\Phi,\mu\nu} \in \[0, \pi\]$.

By \cref{def:kernel-sep},
we have
\begin{align*}
    \Delta_{\min}^{\Phi}  \geq \frac{1}{\beta} \ln{\(\frac{ 2(M-1)}{R_{\Phi}}\)} + 2R_{\Phi},
\end{align*}

and 
\begin{align*}
    \Delta_{\min}^{\Phi} = \Min_{1\leq\mu\leq\nu\leq M}{ \( 1 - \cos{\theta_{\Phi,\mu\nu}} \)} =  \[ 1 - \cos{\theta_{\min}} \],
\end{align*}

where $\theta_{\min} \coloneqq \Min_{1\leq\mu\leq\nu\leq M}{ \( 1 - \cos{\theta_{\Phi, \mu\nu}} \)} \in [0,\pi]$.

Then, it holds
\begin{equation}
\label{eqn:delta-min-lower}
    \[ 1 - \cos{\theta_{\min}} \] \geq \frac{1}{\beta} \ln{\(\frac{ 2(M-1)}{R_{\Phi}}\)} + 2R_{\Phi}.
\end{equation}

Next, Let $1-p$ be the success storage and retrieval probability under \cref{def:kernel-sep}.
We have
\begin{align*}
    P \(  \Delta_{\mu}^{\Phi}   \geq   \frac{1}{\beta} \ln{ \(\frac{ 2(M-1)}{R_{\Phi}}\)} + 2R_{\Phi}   \) = 1 - p.
\end{align*}

By \eqref{eqn:delta-min-lower}, we have
\begin{equation}
\label{eqn:proof0}
    P \(  1 - \cos{\theta_{\min}}    \geq   \frac{1}{\beta} \ln{\(\frac{ 2(M-1)}{R_{\Phi, \mu}}\)} + 2R_{\Phi}   \) = 1 - p.
\end{equation}
We observe that $\cos{\theta_{\min}}$ connects to
the maximal separation loss via
\begin{align*}
    \cos{ \theta_{\min} }  =  \frac{\calL_\Phi( \Xi ) + 2t}{2t}.
\end{align*}

Following the proof of \citep[Lemma~3.1]{hu2023SparseHopfield} and by \cref{lem:technical1}, 
 it holds
\begin{align*}
    M = \sqrt{p} C^{\nicefrac{d-1}{4}},
\end{align*}
with some real value $C \in \R$.
Here 
$C$ is solution to  the upper branch of the Lambert $W$ function deduced from \eqref{eqn:proof0}, 
\begin{align}
C = \frac{b}{ W_0 ( \exp{a + \ln b} ) },
\end{align}
where
\begin{equation}\label{eqn:for-C}
    a \coloneqq \frac{4}{d-1} \left\{  \ln \[ \frac{2( \sqrt{p} - 1 )}{R_\Phi}\] + 1 \right\},
    \quad\text{and}\quad
    b \coloneqq \frac{4 \beta}{5(d-1)}.
\end{equation}

Then, we arrive a lower bound on the exponential storage capacity M:
\begin{align}
M \geq \sqrt{p} C^{\frac{d-1}{4}}.
\end{align}

To compare the the results from \citep[Theorem~3]{ramsauer2020hopfield} (with the assumption of pattern normalization), we denote the results from \cite{ramsauer2020hopfield} with $\tilde{\cdot}$ notation, i.e.
\begin{align*}
    \tilde{a} \coloneqq  \frac{4}{d-1} \left\{  \ln \[ \frac{2( \sqrt{p} - 1 )}{R}\] + 1 \right\},\quad\text{and}\quad
    \tilde{b} = b.
\end{align*}

And we also have $\tilde{\theta}_{\min} \coloneqq \Min_{1\leq\mu\leq\nu\leq M}{ \( 1 - \cos{\theta_{\mu\nu}} \)} \in [0,\pi]$ be the angle between two raw memory patterns $\bxi_\nu, \bxi_\mu$.

We denote the optimal separation loss be $\calL^\star(\Xi)$, and the loss value at $t$-th step be $\calL^t(\Xi)$.

We denote $R_\Phi^\star$ be the corresponding $R_\Phi$ when $\calL_\Phi( \Xi )$ is at its global minimum.

By \eqref{eqn:sep-loss}, 
the convexity of $\calL_\Phi$,
the optimality of $\calL_\Phi$ gives 
\begin{align*}
    R_\Phi^\star =
    \frac{1}{2}
    \sqrt{\frac{ \calL^\star_\Phi(\Xi)}{-t}}
    \ge R_\Phi.
\end{align*}

Next, we prove that to achieve $R_\Phi \geq R$, we need $\calO \( \frac{1}{ -4tR^2 - \calL^\star_{\Phi}( \Xi)}\)$ sub linear time (iterations.).

 Recall that $R  \coloneqq \frac{1}{2} \Min_{\nu,\nu\neq\mu}{\norm{ \bxi_\nu - \bxi_\mu }} $. 
By $R_\Phi=  \sqrt{ \nicefrac{\calL_\Phi(\Xi)}{-t}}/2$,
for $R_\Phi \geq R$, we need
\begin{align*}
    \frac{1}{2}
    \sqrt{\frac{ \calL_\Phi(\Xi)}{-t}}
    \geq
    \frac{1}{2} \Min_{\nu,\nu\neq\mu}{\norm{ \bxi_\nu - \bxi_\mu }} ,
\end{align*}
which implies, by $t>0$, 
\bea
\calL_\Phi(\Xi)
\leq
-t \cdot (2R)^2.
\eea

Subtracting $-\calL^\star_\Phi( \Xi)$ on both sides,
we get
\begin{align*}
\calL_\Phi(\Xi) -\calL^\star_\Phi( \Xi)
&\leq
-t \cdot (2R)^2 -\calL^\star_\Phi( \Xi) \coloneqq \epsilon
\end{align*}

Which implies the iteration number needed to achieve improved memory capacity bound is:
\begin{align*}
     N = \calO \( \frac{1}{ -4tR^2 - \calL^\star_{\Phi}( \Xi)}\),
\end{align*}
which gives us a sub-linear time complexity.

Let 
\begin{align*}
    a \coloneqq \frac{4}{d-1} \left\{  \ln \[ \frac{2( \sqrt{p} - 1 )}{R_\Phi}\] + 1 \right\},
    \quad\text{and}\quad
    b \coloneqq \frac{4 \beta}{5(d-1)}.
\end{align*}

As long as \cref{algorithm1} runs for $\calO \( \nicefrac{1}{ -4tR^2 - \calL^\star_{\Phi}( \Xi)}\)$ iterations, 
its output $\Phi$ satisfies
\begin{align}
\tilde{a} \leq a,
\end{align}
and
\begin{align}
\tilde{C} =   W_0 \( \exp{ \tilde{a} + \ln \tilde{b} } \) \leq W_0 \( \exp{ a + \ln b } \) = C.
\end{align}

Thus, we have the memory capacity comparison as
\begin{align*}
    M = \sqrt{p} C^{\frac{d-1}{4}} \geq \sqrt{p} \tilde{C}^{\frac{d-1}{4}} = \tilde{M}.
\end{align*}

Since the upper branch of  Lambert W function is monotonically increasing on its domain,
$R_\Phi>R$ implies 
\begin{align*}
    M = \sqrt{p} C^{\frac{d-1}{4}} \geq \sqrt{p} \tilde{C}^{\frac{d-1}{4}} = \tilde{M}.
\end{align*}

Hence we finish the proof.
\end{proof}

\clearpage
\subsection{Proofs of \texorpdfstring{\cref{thm:optimal-code-capacity}}{} and \texorpdfstring{\cref{prop:capacity-bound}}{} }\label{proof-of-optimal-code-capacity}

\begin{lemma}[Capacity of Optimal Spherical Code,  \cref{thm:optimal-code-capacity} Restated]
    Given a fixed $D_\Phi > 1$, and its corresponding $M^\star$,
    if an optimal code $\calC_{\text{opt}}$ is in $\mathbb{S}^{D_\Phi-1}$ and has size $M^\star$,
    then $\calC_{\text{opt}} \in \Lambda_{D_\Phi}$.
\end{lemma}

\begin{proposition}[Optimal Memory Capacity, \cref{prop:capacity-bound} Restated]
    Following \cref{thm:optimal-code-capacity},  we have
    \begin{align*}
    M^\star \asymp c^{D_\Phi},     
    \end{align*}
    for some $c>1$.
    Here $\asymp$ indicates matching upper and lower bounds up to constant factors.
\end{proposition}

\begin{proof}
    By \cref{assumption1}, all memories are normalized.
    Thus, we have
    \begin{align}
         R_\Phi 
        &= 
        \frac{1}{2}
        \sqrt{2 - 2\underset{ \underset{\mu\neq\nu}{\mu,\nu\in[M]}}{\max} \Braket{ \Phi(\bxi_\mu), \Phi(\bxi_\nu)}}
        \annot{By \eqref{eq:min_sep}}
        \\
        &=
        \sqrt{\half\Delta_{\min}^\Phi}.
        \label{plug1}
    \end{align}

    Recall the storage condition
    \begin{align*}
        \Delta_{\mu}^\Phi  \geq \frac{1}{\beta} \ln \(  \frac{2(M-1)}{R_\Phi} \).
    \end{align*}

    Here we consider the minimal $\Delta_\mu^\Phi$ among all possible $\mu \in [M]$.
    We plug \eqref{plug1} into the well-separation condition and change $\Delta_\mu^\Phi$ to $\Delta_{\min}^\Phi$.
    We arrive
    \begin{align*}
        \Delta_{\min}^\Phi \geq \frac{1}{\beta} \ln \(  \frac{2(M-1)}{ \sqrt{\Delta_{\min}^\Phi/2 } } \).
    \end{align*}
    By rearranging terms, we get
    \begin{align*}
        \Delta_{\min}^\Phi + \frac{1}{2\beta} \ln \( \frac{1}{2} \Delta_{\min}^\Phi \) \geq 
         \frac{1}{\beta} \ln\( 2(M-1) \).
    \end{align*}
    
    The derivative w.r.t. $\Delta_{\min}^\Phi$ on the LHS is 
    \begin{align*}
        1 + \frac{1}{2\beta \Delta_{\min}^\Phi},
    \end{align*}
    indicating that LHS is increasing in $\Delta_{\min}^\Phi$ for all $\Delta_{\min}^\Phi > 0$.
    The derivative w.r.t. $M$ on the RHS is
    \begin{align*}
        \frac{1}{\beta (M-1)},
    \end{align*}
    indicating that the RHS is increasing in $M$ for all $M > 1$.
    Since we are handling $\Delta_{\min}^\Phi$, this property holds for all $\Delta_\mu^\Phi$.

    Let $\delta$ be the minimum value for $\Delta_{\min}^\Phi$ that satisfies the storage condition such that
    \begin{align*}
        \delta + \frac{1}{2\beta} \ln \( \frac{1}{2} \delta \) \geq 
         \frac{1}{\beta} \ln\( 2(M-1) \).
    \end{align*}
    With the definition of optimal spherical code, we have $\delta \leq 1 - \rho^\star$.
    Thus an optimal spherical code must satisfy this inequality.

    Now we further analyze the quantity $M^\star$.
    Let $\theta =  \arccos\(\rho( C_{\rm opt} )\)$, we apply the upper bound in \cite{kabatiansky1978bounds}, we get
    \begin{align*}
        e^{\varphi(\theta) D_\Phi(1 + o(1))}
        \geq 
        M^\star.
    \end{align*}
    where $\Phi \in \calH$.
    With the above result, we get $M^\star = o(c^{D_\Phi})$ for some $c>1$.
    
    For the lower bound of $M^\star$, we use the classic sphere code bound in \cite{chabauty1953resultats, shannon1959probability, wyner1965capabilities}, and get
    \begin{align*}
        M^\star
        \geq
        \[
        \frac{1}{\sqrt{\pi}}
        \frac{\Gamma( \nicefrac{D_\Phi}{2} )}{ \Gamma( \nicefrac{D_\Phi-1}{2})}
        \int^\theta_0
        \sin^{D_\Phi-2} x
        d x
        \]^{-1}
        =
        (1 + o(1)) \sqrt{2 \pi D_\Phi} \cdot 
        \frac{ \cos{\theta} }{ \sin^{D_\Phi-1}{ \theta} },
    \end{align*}
    where $x \in \mathbb{S}^{D_\Phi - 1}$.

    Therefore, we have
    \begin{align*}
        e^{ \varphi( \theta ) D_\Phi (1 + o(1)) } \
        \geq 
        M^\star
        \geq
        (1 + o(1)) \sqrt{2 \pi D_\Phi} \cdot 
        \frac{ \cos{\theta} }{ \sin^{D_\Phi-1}{ \theta} },
    \end{align*}
    where $\varphi(\theta) > -\log \sin{\theta}$, and $o(\cdot)$ is `` strictly slower than'' notation as $D_\Phi \rightarrow \infty$.

    This completes the proof. Tighter bounds can be found in \cite{jenssen2018kissing, fernandez2021new}. We selected bounds that most clearly show the exponential scaling behavior for better intuition.
\end{proof}

\clearpage
\subsection{Proof of \texorpdfstring{\cref{gamma-conv-2}}{}}\label{proof:gamma-conv-2}

We first restate \cref{gamma-conv-2}:
\begin{theorem}[\cref{gamma-conv-2} Restated]
    For any possible integer $M$, we have 
    \begin{align*}
        \underset{\tau\rightarrow0}{\lim\sup}   
        \(
        \argmin_{ \Phi \in \calH }
        \frac{1}{M} \sum_{\mu=1}^M \calL_\Phi( \xi_\mu, \tau ) \) \subseteq
        \argmin_{ \Phi \in \calH } \calL_{\text{HardMax}}(\Phi)
        ,
    \end{align*}
    where all $\calH$ is the hypothesis space of $\Phi$.
\end{theorem}
Then we introduce a helper lemma.
\begin{lemma}\label{gamma-conv-1}
    Let $\calL_0(\Phi, \tau)$ be 
    \begin{align*}
    \calL_0 \( \Phi, \tau \) \coloneqq \tau \cdot \log \sum_{\mu=1} ^M \calL_\Phi( \xi_\mu, \tau ).
    \end{align*}
    $\calL_0(\Phi, \tau)$ converges uniformly to $\calL_{\text{HardMax}}(\Phi)$ as $\tau \rightarrow 0$.
\end{lemma}

\begin{proof}[Proof of \cref{gamma-conv-2}]

    We first organize terms in \eqref{eqn:sep-loss}. 
    We obtain:
    \begin{align*}
        \calL_\Phi(  \xi_\mu, \tau) 
        &= 
        -  \[
        \log \(
        \exp{\frac{\langle \Phi(\xi_\mu), \Phi(\xi_\mu) \rangle}{\tau}}
        \) 
        - 
        \log
        \(
        \sum_{\nu=1}^M \exp{ \frac{\langle \Phi(\xi_\mu), \Phi(\xi_\nu) \rangle}{\tau}}
        \)
        \]
        \\
        &=
        -  \[
        \frac{1}{\tau}
        - 
        \log
        \(
        \sum_{\nu=1}^M \exp{ \frac{\langle \xi_\mu, \xi_\nu \rangle}{\tau}}
        \)
        \].
    \end{align*}

    We define a helper function $\calL_0$, denoted as
    \begin{align}\label{eqn:helper}
        \calL_0 \( \Phi, \tau \) \coloneqq \tau \cdot \log \sum_{\mu=1} ^M \ell_\mu( \Xi, \Phi, \tau ).
    \end{align}
    
    We have
    \begin{align*}
        \calL_0 \( \Phi, \tau \) 
        &\coloneqq 
        \tau \cdot \log \sum_{\mu=1} ^M \calL_\Phi(\xi_\mu, \tau ) \\
        &=
        \tau \log \sum_{\mu=1} ^M \log \(  1 +  \sum_{\nu\in[M] \backslash \mu } ^M
        \exp{ 
        \frac{\left\langle \Phi(\bxi_\nu), \Phi(\bxi_\mu) \right\rangle - 1}{\tau} }
        \).
    \end{align*}

    Due to the fact that $ \nicefrac{x}{(1+x)} \leq \log(1+x) \leq x $ for all $x > -1$, we have:
    \begin{align}
        & ~ \frac{ \sum_{\nu\in[M] \backslash \mu } ^M
        \exp{ 
        \frac{\left\langle \Phi(\bxi_\nu), \Phi(\bxi_\mu)   \right\rangle - 1}{\tau} } }{1 + \sum_{\nu^\prime\in[M] \backslash \mu } ^M
        \exp{ 
        \frac{\left\langle \Phi(\bxi_\nu), \Phi(\bxi_\mu)   \right\rangle - 1}{\tau} }} 
        \notag\\
        \leq & ~ 
        \log \(  1 +  \sum_{\nu\in[M] \setminus \mu } ^M
        \exp{ 
        \frac{\left\langle \Phi(\bxi_\nu), \Phi(\bxi_\mu)   \right\rangle - 1}{\tau} }
        \)
        \notag\\
        \leq & ~ 
        \sum_{\nu\in[M] \backslash \mu } ^M
        \exp{ 
        \frac{\left\langle \Phi(\bxi_\nu), \Phi(\bxi_\mu)   \right\rangle - 1}{\tau} }.
        \label{up-low-lim-gamma}
    \end{align}
    Given the fact that $\left\langle \Phi(\bxi_\nu), \Phi(\bxi_\mu)   \right\rangle - 1 \leq 0$ for all possible $\nu, \mu$.
    With the monotonicity of the exponential function, we obtain:
    \begin{align*}
        \sum_{\nu\in[M] \backslash \mu } ^M
        \exp{ 
        \frac{\left\langle \Phi(\bxi_\nu), \Phi(\bxi_\mu)   \right\rangle - 1}{\tau} }
        \leq M - 1.
    \end{align*}

    Combining this with LHS of \eqref{up-low-lim-gamma}, we have
    \begin{align*}
         & ~ \frac{ \sum_{\nu\in[M] \backslash \mu } ^M
        \exp{ 
        \frac{\left\langle \Phi(\bxi_\nu), \Phi(\bxi_\mu)   \right\rangle - 1}{\tau} } }{M} \\
        \leq  & ~
        \log \(  1 +  \sum_{\nu\in[M] \backslash \mu } ^M
        \exp{ 
        \frac{\left\langle \Phi(\bxi_\nu), \Phi(\bxi_\mu)   \right\rangle - 1}{\tau} }
        \)
        \\
        \leq & ~
        \sum_{\nu\in[M] \backslash \mu } ^M
        \exp{ 
        \frac{\left\langle \Phi(\bxi_\nu), \Phi(\bxi_\mu)   \right\rangle - 1}{\tau} }.
    \end{align*}
    
    Summing over all possible $\mu \in [M]$ we have

    \begin{align*}
    & ~ \sum_{\mu=1}^M
    \frac{ \sum_{\nu\in[M] \backslash \mu } ^M
    \exp{ 
    \frac{\left\langle \Phi(\bxi_\nu), \Phi(\bxi_\mu)   \right\rangle - 1}{\tau} } }{M} \\
    \leq & ~ 
    \sum_{\mu=1}^M
    \log \(  1 +  \sum_{\nu\in[M] \backslash \mu } ^M
    \exp{ 
    \frac{\left\langle \Phi(\bxi_\nu), \Phi(\bxi_\mu)   \right\rangle - 1}{\tau} }
    \)
    \\
    \leq & ~ 
    \sum_{\mu=1}^M
    \sum_{\nu\in[M] \backslash \mu } ^M
    \exp{ 
    \frac{\left\langle \Phi(\bxi_\nu), \Phi(\bxi_\mu)   \right\rangle - 1}{\tau} }.
    \end{align*}

    Using the property of max function, we further get
    \begin{align*}
    & ~  \max_{\mu,\nu\in[M], \mu\neq\nu}
    \frac{ 
    \exp{ 
    \frac{\left\langle \Phi(\bxi_\nu), \Phi(\bxi_\mu)   \right\rangle - 1}{\tau} } }{M} 
    \\
    \leq & ~ 
    \sum_{\mu=1}^M
    \frac{ \sum_{\nu\in[M] \backslash \mu } ^M
    \exp{ 
    \frac{\left\langle \Phi(\bxi_\nu), \Phi(\bxi_\mu)   \right\rangle - 1}{\tau} } }{M} ,
    \\
    \leq & ~ 
    \sum_{\mu=1}^M
    \log \(  1 +  \sum_{\nu\in[M] \backslash \mu } ^M
    \exp{ 
    \frac{\left\langle \Phi(\bxi_\nu), \Phi(\bxi_\mu)   \right\rangle - 1}{\tau} }
    \)
    \\
    \leq & ~ 
    M\cdot(M-1) \cdot
    \max_{\mu,\nu\in[M], \mu\neq\nu}
    \exp{ 
    \frac{\left\langle \Phi(\bxi_\nu), \Phi(\bxi_\mu)   \right\rangle - 1}{\tau} }.
    \end{align*}

    Now by taking logarithmic on both sides and multiplying all three terms by $\tau$ we get
    \begin{align*}
    \max_{\mu,\nu\in[M], \mu\neq\nu}
    \(\left\langle \Phi(\bxi_\nu), \Phi(\bxi_\mu)   \right\rangle - 1 \)
    - \tau \log M 
    \leq
    \calL_0( \Phi, \tau )    
    \leq
    \tau \log\(M\cdot(M-1)\)
    +
    \max_{\mu,\nu\in[M], \mu\neq\nu}
    \(\left\langle \Phi(\bxi_\nu), \Phi(\bxi_\mu)   \right\rangle - 1 \) .
    \end{align*}

    By $\max_{\mu,\nu\in[M], \mu\neq\nu}
    \(\alpha_{ \mu,\nu } \)  = \calL_{\text{HardMax}}(\Phi) - 1$, we have
    \begin{align*}
    \calL_{\text{HardMax}}(\Phi)
    - \tau \log M 
    - 1
    \leq
    \calL_0( \Phi, \tau )    
    \leq
    \tau \log{M\cdot(M-1)}
    +
    \calL_{\text{HardMax}}(\Phi) - 1.
    \end{align*}   

    Therefore, for any $\epsilon > 0$, 
    by taking $\tau_0 = \frac{\epsilon}{ \max{\( \log M, \log (M \cdot (M-1)) \)} }$, 
    we have
    \begin{align*}
        \abs{
        \calL_0(\Phi, \tau) - \calL_{\text{HardMax}}(\Phi)}
        \leq 
        \tau \max\{\log M, \log (M \cdot (M-1))\} \leq \epsilon,
    \end{align*}
    for any $\tau < \tau_0$.
    That is, $\calL_0(\Phi, \tau)$ converges uniformly to $\calL_{\text{HardMax}}(\Phi)$, leading to \cref{gamma-conv-1}.

    Now we know $\calL_0(\Phi, \tau)$ converges uniformly to $\calL_{\text{HardMax}}(\Phi)$ as $\epsilon \rightarrow 0$, by \cite[Proposition~7.15]{rockafellar2009variational}, we have $\calL_0(\Phi, \tau)$ $\Gamma$-converges to $\calL_{\text{HardMax}}(\Phi)$ as well.
    By \cite[Theorem~2.10]{braides2006handbook}, we have
    \begin{align*}
        \liminf_{ \tau \rightarrow 0 }
        \argmin_{\Phi \in \calH}
        \calL_0(\Phi, \tau)
        \subseteq
        \argmin_{\Phi \in \calH}
        \calL_{\text{HardMax}}(\Phi).
    \end{align*}
    This completes the proof\footnote{
    In general, $\calL_0$ converges uniformly to $\calL_{\text{hardmax}}$ as $\tau$ goes to 0 with an error rate of $| \mathcal{L}_0 - \mathcal{L}_{\text{hardmax}}  | \leq 2 \tau \log(M)$.
    }. 
\end{proof}

\subsection{Proof of \texorpdfstring{\cref{lemma:separation-bound}}{}}\label{proof:separation-bound}

\begin{proof}
We first define the one-vs-one distance.

\begin{definition}[one-vs-one distance]
    We define the one-vs-one distance of a set of points $\calV = \{\bv_\mu\}_{\mu=1}^M \subseteq \mathbb{S}^{d-1}$, with $\abs{ \calV } = M$, as
    \begin{align*}
        \rho_{\text{one-vs-one}} \(  \calV \)
        \coloneqq
        \underset{ \mu\in[M]}{\min} \underset{\nu\neq\mu}{\min}
        \norm{ \bv_\nu - \bv_\mu  }.
    \end{align*}
\end{definition}

The one-vs-one distance is lower bounded as following
\begin{lemma}{\citep[Lemma~C.13]{jiang2023generalized}
}\label{lem:gnc}
\begin{equation}
    \[  \frac{\sqrt{\pi}}{M}   \frac{ \Gamma \( \frac{d+1}{2} \) }{  \Gamma \( \frac{d}{2} + 1 \) }  \]^{\frac{1}{d-1}}
    \leq
    \underset{\calV \subseteq \mathbb{S}^{d-1}}{\max}
    \rho_{\text{one-vs-one}} \(  \calV \).
    \end{equation}
\end{lemma}

Note that by the definition of the one-vs-one distance, we have the equivalent expression such that
\begin{align*}
    \frac{\rho^2_{\text{one-vs-one}} \(  \calV \)}{2}
    \equiv
    \min_{\nu,\mu\in[M]} \Braket{ \bv_\mu, \bv_\mu } - \Braket{\bv_\mu, \bv_\nu}.
\end{align*}
Combining the above property, \cref{lem:gnc} and a known upper bound in \citep[Theorem~1]{moore1974vector}, we obtain:
\begin{align*}
    \frac{1}{2}
    \[ 
    \frac{\sqrt{\pi}}{M}
    \frac{\Gamma\( \frac{D_\Phi+1}{2} \)  }{ \Gamma\( \frac{D_\Phi}{2} + 1 \) }
    \]^{ \frac{2}{D_\Phi-1}  }
    \leq
    \Delta_{\min}^\Phi
    \leq
    2
    \[
    \frac{2\sqrt{\pi}}{M}
    \frac{ \Gamma \(  \frac{D_\Phi+1}{2} \) }{  \Gamma \(  \frac{D_\Phi}{2}  \) }
    \]^{\frac{1}{D_\Phi-1}}.
\end{align*}
The upper bound in \cite{moore1974vector} is derived by the normalized surface area of a spherical cap of angular radius $\theta$.
\end{proof}

\clearpage

\section{Experimental Details}\label{appendix:exp-details}

\paragraph{Computational Environments.}
All experiments are conducted on the platform with NVIDIA GEFORCE RTX 2080 Ti and INTEL XEON SILVER 4214 @ 2.20GHz.
We use PyTorch 1.8.0 for all experiments.
The experiments are relatively lightweight which can also be ran on CPU-only environments.

\subsection{Metastable States}\label{appendix:meta-detail}

\paragraph{Hyperparameters.}

The hyperparameters we used for the metastable state experiment is listed in \cref{table:hyper-meta}.

\begin{table}[h]
        \centering
        \caption{Hyperparameter used Metastable State Experiment.
        }
        \begin{tabular}{l*{2}{c}}
        \toprule
            \bf{parameter} & \multicolumn{1}{c}{\bf{Synthetic}} & \multicolumn{1}{c}{\bf{MNIST}}\\ 
            \midrule
            Optimizer & Adam & Adam \\
            Learning Iteration $N$ & 20 & 20 \\
            Batch Size & 10 & 16 \\
            Update Rule Iteration & 20 & 5 \\
            Learning Rate  & $0.1$ & $0.1$ \\
            Memory set size $M$ & $10$ & $60000$ \\
            Pattern Dimension $d$ & $5$ & $784$ \\
            Feature Dimension $D_\Phi$ & $5$ & $200$ \\
            $\beta$ & $4$ & $0.1$ \\
            threshold for $\bp$ & $0.01$ & $0.01$ \\
            \bottomrule
        \end{tabular}
        \label{table:hyper-meta}
    \end{table} 

\paragraph{Implementation Details.}
The batch size in \cref{table:hyper-meta} denotes the batch size we use to train the feature map $\Phi$.
For the synthetic dataset, we directly train $\Phi$ on the whole memory set.
For the softmax threshold, we follow the settings used in \cite{santos2024sparse}.

\subsection{Energy Landscape}\label{appendix:landscape-detail}
\paragraph{Hyperparameters.}

The hyperparameters we used for the basins of attraction experiment is listed in \cref{table:basins}.

\begin{table}[h]
        \centering
        \caption{Hyperparameter used in the energy landscape experiment.
        }
        \begin{tabular}{l*{2}{c}}
        \toprule
            \bf{parameter} & \multicolumn{1}{c}{\bf{2-Points}} & \multicolumn{1}{c}{\bf{4-Points}} \\ 
            \midrule
            Optimizer & SGD & SGD  \\
            Learning Iteration $N$ & 5 & 5 \\
            Learning Rate  & $0.1$ & $0.1$ \\
            Memory set size $M$ & $2$ & $4$  \\
            Pattern Dimension $d$ & $2$ & $2$ \\
            Feature Dimension $D_\Phi$ & $2$ & $2$  \\
            $\beta$ & $20$ & $0.9^{-1}$ \\
            query grid resolution & $40\times 40$ & $40\times 40$ \\
            color map & hot & hot \\
            \bottomrule
        \end{tabular}
        \label{table:contours}
    \end{table} 

\paragraph{Implementation Details.}
We first prepare a set of randomly generated patterns as memories.
Next we record its energy landscape with respect to different query (the coordinate in the figure).
Next we train $\Phi$ for 5 iterations and record its resulting energy landscape with $N=1, 2, 5$.
We use the entmax and sparsemax package used in \cite{peters2019sparse}.

\clearpage
\subsection{Basins of Attraction}\label{appendix:basins}

\paragraph{Hyperparameters.}

The hyperparameters we used for the basins of attraction experiment is listed in \cref{table:basins}.

\begin{table}[h]
        \centering
        \caption{Hyperparameter used in the basins of attraction experiment.
        }
        \begin{tabular}{l*{2}{c}}
        \toprule
            \bf{parameter} & \multicolumn{1}{c}{\bf{Synthetic}}  \\ 
            \midrule
            Optimizer & Adam  \\
            Learning Iteration $N$ & 5 \\
            Update Rule Iteration & 5 \\
            Learning Rate  & $0.1$ \\
            Memory set size $M$ & $5$  \\
            Pattern Dimension $d$ & $5$ \\
            Feature Dimension $D_\Phi$ & $5$  \\
            $\beta$ & $20$  \\
            query grid resolution & $100\times 100$ \\
            \bottomrule
        \end{tabular}
        \label{table:basins}
    \end{table} 

\paragraph{Implementation Details.}
We specifically set the update rule iteration to $5$ as we see the sharp energy gradient in \cref{fig:contours}.
Demonstrating that the standard MHM and its variants are not able to converge to fixed points fast.

\subsection{Simulation of \texorpdfstring{\cref{lemma:separation-bound}}{}}\label{separation-visualize}

We provide a numerical simulation of our bound with $D_\Phi = 3$.
We take the known solution of minimal separation published in \url{http://neilsloane.com/packings/} a ground truth.
\begin{figure}[!h]
    \centering
    \includegraphics[width=0.7\textwidth]{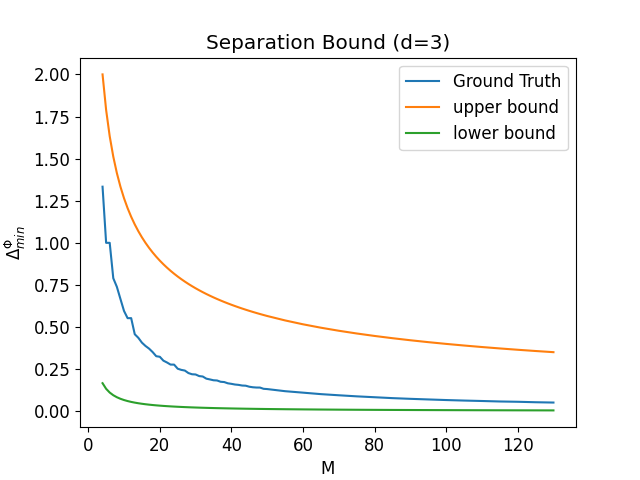}
    \vspace{-0.5em}
    \caption{
    \textbf{Separation Bound Numerical Simulation}
    We visualize the bound presented in \cref{lemma:separation-bound} in 3-D dimension.
    The bound goes tighter as the number of points increases.
    }
    \label{fig:sep-bound}
\end{figure}

\subsection{Assignment Problems}\label{appendix:assignment}

Here we conduct the point assignment problem in 2D space.
In 2D space, the optimal arrangement of six points is well-established: they equally divide the unit circle, with each point neighboring two others. 

\begin{figure}[!h]
    \centering
    \includegraphics[width=0.7\textwidth]{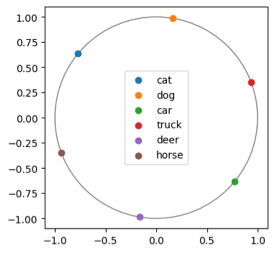}
    \caption{\textbf{Assignment Problem in 2D}
    We observe that the learned feature map consistently put similar pairs closer to each other, leading to preserving some level of semantic information.
    }
    \label{fig:assignment}
\end{figure}

We consider the case where we try to learn a feature map $\Phi$ under \uhop+ with 6 different images sampled from CIFAR10
We sample each image from cat, dog, car, truck, deer and horse class.
Intuitively, cat has closer semantic relationship with dog, car is more similar to truck and deer has closer semantic relationship with horse \cite{jiang2023generalized, neelakantan2022text}.
We show that our learned feature map consistently puts similar pairs closer to each other in 7 out of 10 trials.
This implies that while our method does not force or even considers the underlying semantic meanings behind each memories, our feature map is still able to present such relationship.
The result is in \cref{fig:assignment}.
\paragraph{Discussion.}

The separation loss encourages the feature map to make the whole dataset the most linearly separable to each instance.
A similar analysis can be found in \cite{JMLR:v23:21-1079, ghosal2022randomly}, where they found out that if data has subtle clustered structure, a random neural network is able to make it linearly separable with high probability. 

\subsection{Additional Experiments}
Here we observe the loss curve of $\calL$ w.r.t. different memory set size.
We aim to verify whether $\calL$ is able to converge well through proposed algorithm.

\begin{figure*}[b]
\centering
\includegraphics[width=0.7\linewidth]{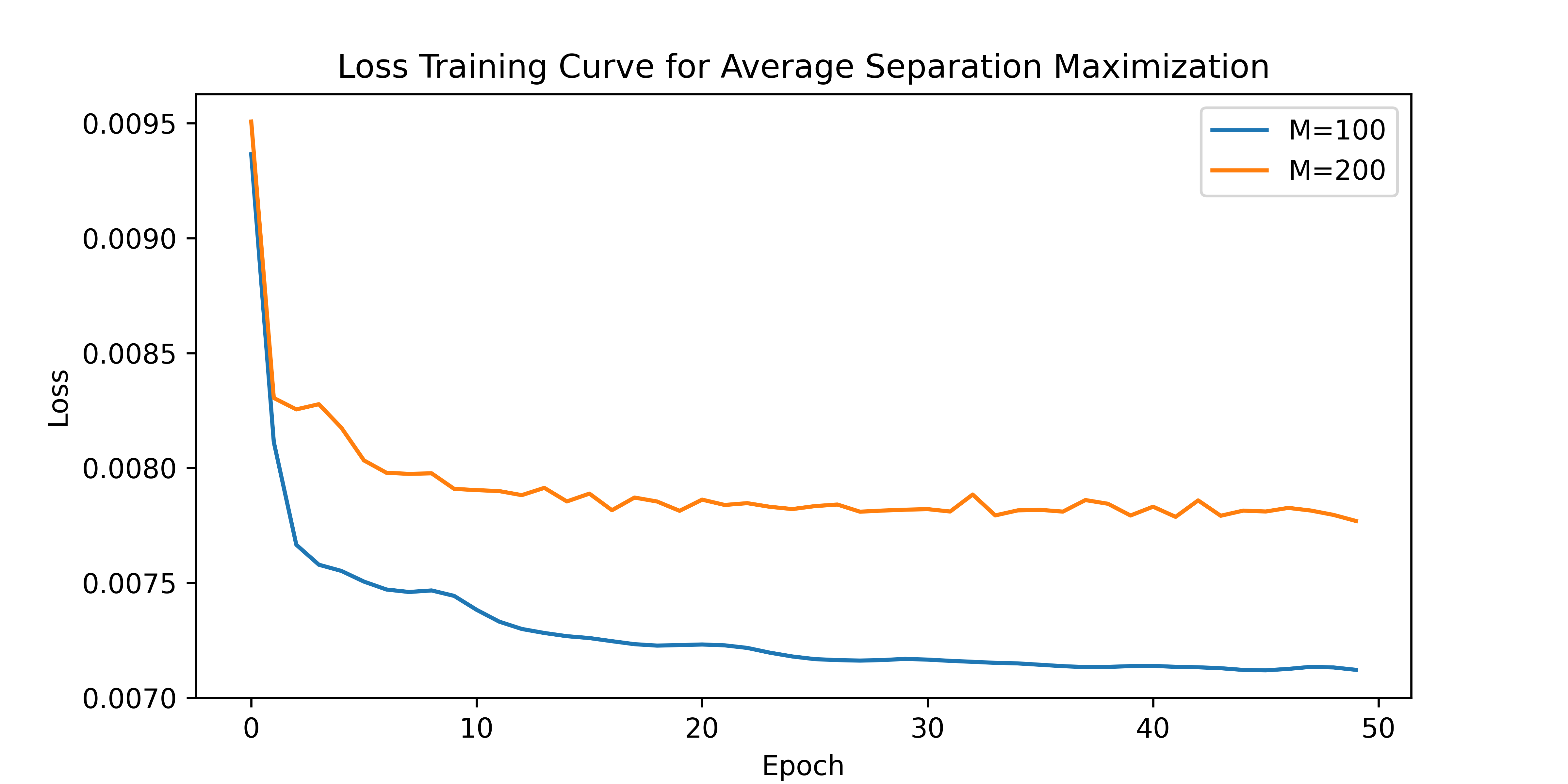}
\caption{\textbf{Loss Curve of $\calL$ w.r.t. different memory set size}. 
We run separation maximization for 100 epochs on MNIST under 2 settings, $M=100/200$.
We set $\tau=0.1$, learning rate 1e-3, $D_\Phi=100$.
The result shows $\calL$ converges fast, which echoes our sub-linear time complexity. 
}
\label{fig:loss-conv}
\end{figure*}

\clearpage
\newpage
\setlist[itemize]{leftmargin=1em}
\setlist[enumerate]{leftmargin=1.4em}
\section*{NeurIPS Paper Checklist}

\begin{enumerate}

\item {\bf Claims}
    \item[] Question: Do the main claims made in the abstract and introduction accurately reflect the paper's contributions and scope?
    \item[] Answer: \answerYes{} %
    \item[] Justification: 
    In our abstract, we claim we propose a new framework to describe the optimal memory capacity of kernelized Hopfield models (KHMs) and modern Hopfield models.
    We consdier stored memory sets as a special type of spherical code that all points in the set satisfies the well-separation condition.
    Next, we show there is a sublinear time algorithm to find an optimal feature map for KHMs to achieve maximal memory capacity.
    The main claims are detailed described in the following sections
    \begin{itemize}
        \item 
        Optimal memory capacity for KHMs and MHMs:
        \cref{sec:method}, \cref{thm:optimal-code-capacity}.
        \item
        Memory code:
        \cref{method:spherical-code}, \cref{def:memory-code}.
        \item
        Optimal capacity algorithm: \cref{method:algorithm}, \cref{algorithm1}.
    \end{itemize}
    

\item {\bf Limitations}
    \item[] Question: Does the paper discuss the limitations of the work performed by the authors?
    \item[] Answer: \answerYes{} %
    \item[] Justification: 
    We discussed our limitations in the last section, the \textbf{limitations} paragraph.
    The time complexity of \cref{algorithm1} was discussed in \cref{method:algorithm}.

\item {\bf Theory Assumptions and Proofs}
    \item[] Question: For each theoretical result, does the paper provide the full set of assumptions and a complete (and correct) proof?
    \item[] Answer: \answerYes{} %
    \item[] Justification:
    We have several theoretical results:
    \begin{enumerate}
        \item 
        \cref{lem:KHM-lower-bound}: Proof in \cref{proof:capacity-lower-bound}.
        \item
        \cref{thm:optimal-code-capacity}: Proof in \cref{proof-of-optimal-code-capacity}.
        \item
        \cref{gamma-conv-2}: Proof in \cref{proof:gamma-conv-2}.
        \item
        \cref{lemma:separation-bound}: Proof in \cref{proof:separation-bound}.
    \end{enumerate}
    The main assumption we made is normalized memory patterns, which is described in \cref{lem:KHM-lower-bound} and \cref{thm:optimal-code-capacity}.
    Similar assumption was also made in \cite{santos2024sparse, wu2024uniform}.

    \item {\bf Experimental Result Reproducibility}
    \item[] Question: Does the paper fully disclose all the information needed to reproduce the main experimental results of the paper to the extent that it affects the main claims and/or conclusions of the paper (regardless of whether the code and data are provided or not)?
    \item[] Answer: \answerYes{} %
    \item[] Justification:
    Our experimental results are in \cref{sec:exp}.
    We describe our experimental details in both \cref{sec:exp} and \cref{appendix:exp-details}.

\item {\bf Open access to data and code}
    \item[] Question: Does the paper provide open access to the data and code, with sufficient instructions to faithfully reproduce the main experimental results, as described in supplemental material?
    \item[] Answer: \answerYes{} %
    \item[] Justification: 
    We provide the code in our supplementary materials.
    As for data, we mainly use synthetic data and MNIST in all experiments.
    We also describe the data generation and download in our code and experimental details.

\item {\bf Experimental Setting/Details}
    \item[] Question: Does the paper specify all the training and test details (e.g., data splits, hyperparameters, how they were chosen, type of optimizer, etc.) necessary to understand the results?
    \item[] Answer: \answerYes{} %
    \item[] Justification: 
    The details are fully described in \cref{appendix:exp-details}.

\item {\bf Experiment Statistical Significance}
    \item[] Question: Does the paper report error bars suitably and correctly defined or other appropriate information about the statistical significance of the experiments?
    \item[] Answer: \answerYes{} %
    \item[] Justification: 
    The results are ran over at least 5 runs with different random seeds.

\item {\bf Experiments Compute Resources}
    \item[] Question: For each experiment, does the paper provide sufficient information on the computer resources (type of compute workers, memory, time of execution) needed to reproduce the experiments?
    \item[] Answer: \answerYes{} %
    \item[] Justification: 
    We describe our computational environment in the first paragraph of \cref{appendix:exp-details}.
    
\item {\bf Code Of Ethics}
    \item[] Question: Does the research conducted in the paper conform, in every respect, with the NeurIPS Code of Ethics \url{https://neurips.cc/public/EthicsGuidelines}?
    \item[] Answer: \answerYes{} %
    \item[] Justification: 
    The authors have read and agreed to every aspect of the NeurIPS Code of Ethics.

\item {\bf Broader Impacts}
    \item[] Question: Does the paper discuss both potential positive societal impacts and negative societal impacts of the work performed?
    \item[] Answer: \answerYes{} %
    \item[] Justification: 
    The broader impacts can be found in the section right after conclusions.
    
\item {\bf Safeguards}
    \item[] Question: Does the paper describe safeguards that have been put in place for responsible release of data or models that have a high risk for misuse (e.g., pretrained language models, image generators, or scraped datasets)?
    \item[] Answer: \answerNA{} %
    \item[] Justification: 
    The experiments conducted are mostly numerical simulations which does not serve practical usage.

\item {\bf Licenses for existing assets}
    \item[] Question: Are the creators or original owners of assets (e.g., code, data, models), used in the paper, properly credited and are the license and terms of use explicitly mentioned and properly respected?
    \item[] Answer: \answerYes{} %
    \item[] Justification: 
    For our experiments, we only use datasets provided or generated by PyTorch. PyTorch's licenses can be found in \url{https://github.com/pytorch/pytorch/blob/main/LICENSE}.

\item {\bf New Assets}
    \item[] Question: Are new assets introduced in the paper well documented and is the documentation provided alongside the assets?
    \item[] Answer: \answerNA{} %
    \item[] Justification: 
    The paper does not provide new assets.

\item {\bf Crowdsourcing and Research with Human Subjects}
    \item[] Question: For crowdsourcing experiments and research with human subjects, does the paper include the full text of instructions given to participants and screenshots, if applicable, as well as details about compensation (if any)? 
    \item[] Answer: \answerNA{} %
    \item[] Justification: This paper does not involve any crowdsourcing our human subjects experiments.

\item {\bf Institutional Review Board (IRB) Approvals or Equivalent for Research with Human Subjects}
    \item[] Question: Does the paper describe potential risks incurred by study participants, whether such risks were disclosed to the subjects, and whether Institutional Review Board (IRB) approvals (or an equivalent approval/review based on the requirements of your country or institution) were obtained?
    \item[] Answer: \answerNA {} %
    \item[] Justification:  This paper does not involve any crowdsourcing our human subjects experiments.

\end{enumerate}

\end{document}